\documentclass{article} % For LaTeX2e
\usepackage{iclr2025_conference,times}

\usepackage{amsmath,amsfonts,bm}

\def\eqref#1{equation~\ref{#1}}
\def\1{\bm{1}}

\DeclareMathAlphabet{\mathsfit}{\encodingdefault}{\sfdefault}{m}{sl}
\SetMathAlphabet{\mathsfit}{bold}{\encodingdefault}{\sfdefault}{bx}{n}

\def\ours{NavFoM\xspace}
\DeclareUnicodeCharacter{FF0C}{,}

\definecolor{myblue}{HTML}{dbe8f5}
\definecolor{mygreen}{HTML}{009900}

\definecolor{BestColor}{HTML}{ffffff}
\definecolor{SecondColor}{HTML}{ffffff}
\usepackage[table]{xcolor}
\usepackage{titletoc}
\usepackage{tocloft}
\usepackage{lipsum}
\usepackage{url}
\usepackage{graphicx}
\usepackage{subcaption}
\usepackage{listings}
\usepackage[nolist]{acronym}
\usepackage{amsmath}
\usepackage[font={small}]{caption}
\usepackage[export]{adjustbox}
\usepackage{amssymb}
\usepackage{bm}
\usepackage{booktabs}
\usepackage{balance}
\usepackage{color}
\usepackage{dsfont}
\usepackage{esvect}
\usepackage{float}
\usepackage{graphicx}
\usepackage{import}
\usepackage{mathtools}
\usepackage{multirow}
\usepackage{multicol}
\usepackage{flushend}
\usepackage{outline}
\usepackage{paralist}
\usepackage{siunitx}
\usepackage{stackengine}
\usepackage{stmaryrd}
\usepackage{systeme}
\usepackage{soul}
\usepackage{url}
\usepackage{tikz}
\usetikzlibrary{tikzmark}
\usetikzlibrary{calc}
\usepackage[normalem]{ulem}

\usepackage{enumitem}
\usepackage{algorithm2e}
\usepackage{scalerel}
\usepackage{makecell}
\usepackage{wrapfig}

\definecolor{citecolor}{HTML}{0071bc}
\usepackage[pagebackref=true,breaklinks=true,urlcolor=blue,colorlinks,citecolor=citecolor,bookmarks=false]{hyperref}

\title{\quad \; Embodied Navigation Foundation Model}

\author{%
    \begin{minipage}{\linewidth}
    \centering
    \vspace{8pt}
	Jiazhao Zhang$^{1,2,*}$ ~
	Anqi Li$^{1,2,*}$ ~
	Yunpeng Qi$^{3,4,*}$ ~
	Minghan Li$^{2,*}$ ~
    Jiahang Liu$^2$ ~\\
    \textbf{Shaoan Wang$^1$ ~
    Haoran Liu$^{1,2}$ ~
    Gengze Zhou$^5$ ~
    Yuze Wu$^6$ ~
    Xingxing Li$^6$ ~
    Yuxin Fan$^6$ ~}\\
    \textbf{Wenjun Li$^6$~
    Zhibo Chen$^3$ ~
    Fei Gao$^{6,7}$ ~
    Qi Wu$^5$ ~
    Zhizheng Zhang$^{2,4,\dagger} ~$
    He Wang$^{1,2,4,\dagger}$ }
    \end{minipage}
	\\\\
    \begin{minipage}{\linewidth}
    \centering
    \begin{tabular}{c}
    $^1$Peking University~
	$^2$Galbot~
    $^3$USTC~
    $^4$BAAI \\
    $^5$University of Adelaide~
    $^6$Zhejiang University
    $^7$Differential Robotics
    \end{tabular}
    \end{minipage}
    \\
    \begin{minipage}{\linewidth}
    \centering
    \begin{tabular}{c}
    Project Page: \url{https://pku-epic.github.io/NavFoM-Web/}
    \end{tabular}
    \end{minipage}
}
\iclrfinalcopy % Uncomment for camera-ready version, but NOT for submission.
\begin{document}
\def\thefootnote{$^*$}\footnotetext{Joint First Author\quad $^\dagger$ Corresponding Author}

\maketitle
\begin{abstract}
% Navigation is a fundamental capability in embodied AI, representing the intelligence required to perceive and interact within physical environments, To statify the reuqired inteignece, recent advaced works leverage a similar foumration with strong generabailtiy technical, Vision-Language Models (VLMs), but remains largely confined to narrow task settings and embodiment-specific architectures. 

Navigation is a fundamental capability in embodied AI, representing the intelligence required to perceive and interact within physical environments. To achieve such intelligence, recent advanced works leverage Vision-Language Models (VLMs), which demonstrate strong generalizability and possess a well-suited formulation for navigation. However, these approaches remain largely confined to narrow task settings and embodiment-specific architectures.
In this work, we introduce a cross-embodiment and cross-task Navigation Foundation Model (\ours), trained on eight million navigation samples that encompass quadrupeds, drones, wheeled robots, and vehicles, and spanning diverse tasks such as vision-and-language navigation, object searching, target tracking, and autonomous driving.
\ours employs a unified architecture that processes multimodal navigation inputs from varying camera configurations and navigation horizons.
To accommodate diverse camera setups and temporal horizons, \ours incorporates identifier tokens that embed camera view information of embodiments and the temporal context of tasks. 
Furthermore, to meet the demands of real-world deployment, \ours controls all observation tokens using a dynamically adjusted sampling strategy under a limited token length budget.
Extensive evaluations on seven public benchmarks demonstrate that our model achieves state-of-the-art or highly competitive performance across different navigation tasks and embodiments without requiring task-specific fine-tuning. Additional real-world experiments further confirm the strong generalizability and practical applicability of our approach.
\end{abstract}

\begin{figure}[ht]
  \centering
    \includegraphics[width=0.89\linewidth]{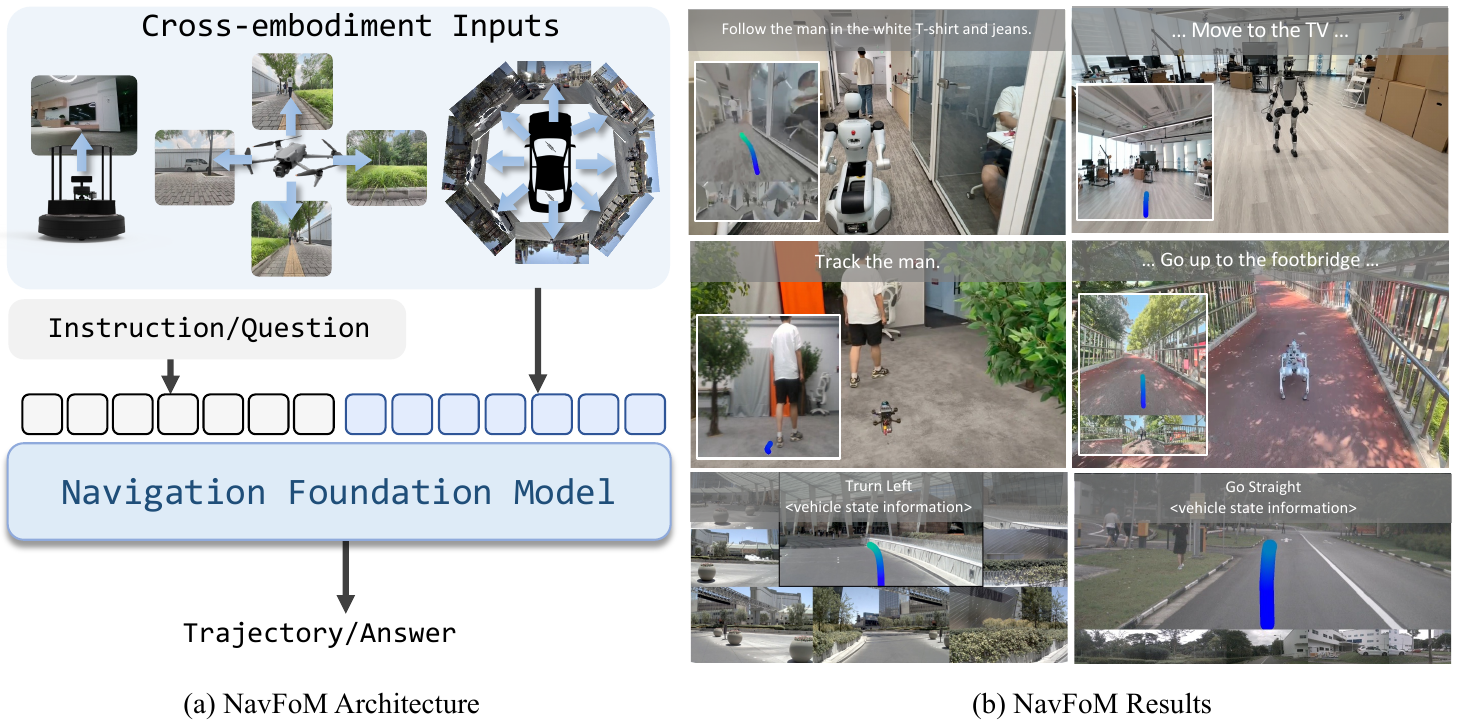}
  \caption{We provide an illustration of architecture (left) alongside real-world experiment results (right). The results include cross-task and cross-embodiment performence of our method.}
  \label{fig:teaser}
\end{figure}

\section{Introduction}
For both embodied agents and humans, navigation serves as a foundational capability that enables them to move intelligently within physical environments to accomplish specified tasks~\citep{shah2023gnm, bar2025navigation, Zhang2024VisionandLanguageNT}. Achieving robust navigation requires a deep understanding of environmental context and task instructions, typically presented through visual and linguistic observations, which are reminiscent of Visual Language Models (VLMs). However, VLMs~\citep{liu2023llava, qwen2,guo2025seed1} have recently demonstrated remarkable zero-shot generalization in tasks such as retrieval, classification, and captioning from large-scale open-world data, without reliance on domain-specific fine-tuning. In contrast, embodied navigation~\citep{habitat19iccv, deitke2022procthor} remains tied to narrow task domains, embodiment-specific architectures, and restricted instruction formats.

In pursuit of generalist navigation, the community has witnessed growing interest~\citep{zhang2024navid, cheng2024navila, shah2023gnm,long2024instructnav}, yet progress has been hindered by the constrained design and limited domain applicability of prior research. In cross-task navigation, previous methods~\citep{zhang2024uni, yin2025unigoal, zhu2025mtu} typically assume a consistent camera configuration for the robot and unify various tasks such as vision-and-language navigation, object searching, and target tracking. For cross-embodiment navigation, current approaches~\citep{eftekhar2024one, hirose2023exaug} implicitly learn priors about the physical shape of the embodiment but are often restricted to specific navigation tasks. The existing divergence between navigation tasks and embodiments highlights the absence of a foundational navigation model capable of handling different tasks across diverse embodiments.

In this work, we toward building a cross-task and cross-embodiment embodied navigation foundation model, \ours{}, trained on eight million navigation samples spanning diverse embodiments and tasks. Inspired by humans’ ability to accomplish a wide range of navigation tasks primarily through visual sensory input and the recent success of vision-only navigation methods~\citep{shah2023gnm, zeng2024poliformer}, we formulate the generalist navigation task as processing egocentric videos (captured by one or more cameras mounted on the robot) alongside language instructions, and predicting subsequent trajectories to fulfill those instructions. This formulation is compatible with most existing navigation task settings~\citep{openscene2023, wang2024openuav}.

% \gz{``Inspired by humans’ ability...", shall we mention other robotic systems that integrate other sensors to justify our design choice? e.g. ``visual sensory input is scalable and enough"}

To align generalizable embodiments across diverse camera configurations, we introduce temporal-viewpoint indicator tokens (TVI tokens) to identify both the viewpoint of camera setups and the temporal information of the navigation horizon. By dynamically adjusting these TVI tokens, our method enables co-tuning across different camera setups and supports joint training with both image-QA and video-QA samples~\citep{shen2024longvu, li2023llama}. Furthermore, to address the constraints of practical deployment such as hardware memory cost and inference speed, we propose a token Budget-Aware Temporal Sampling (BATS) strategy, which dynamically samples navigation history tokens based on a forgetting curve constrained by a token budget. This token sampling approach balances performance and inference speed, enhancing the practicality of our method for real-world deployment.

\begin{figure}[t]
  \centering
    \includegraphics[width=\linewidth]{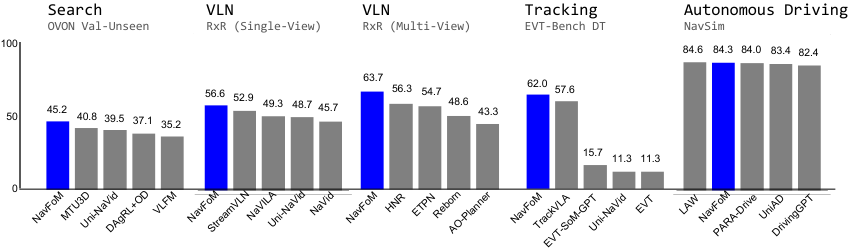}
  \caption{Benchmark performance of \ours{}, we compare \ours{} with SOTA baselines on each benchmarks. See Sec.~\ref{sec:exp} for more detials.}
  \label{fig:teaser_comp}
\end{figure}

We collected a comprehensive and diverse navigation dataset comprising 8.02 million samples, sourced from public navigation datasets~\citep{habitat19iccv,wang2025trackvla,openscene2023,wang2024openuav} and pseudo web-video navigation data~\citep{li2025sekai}. The dataset includes cross-embodiment trajectories from quadruped robots, drones, wheeled robots, and cars, covering a wide range of tasks such as vision-and-language navigation, object searching, target tracking, and autonomous driving. These navigation samples feature diverse instructions and scenarios that require multiple skills, enabling \ours{} to acquire generalized navigation capabilities.
Additionally, we gathered 4.76 million open-world knowledge samples~\citep{shen2024longvu, li2023llama} derived from both image-based and video-based question-answering tasks. Following the approach of \citep{zhang2024navid}, we co-tune the navigation data together with image and video QA data in an end-to-end manner, facilitating large-scale and comprehensive training of \ours{}.

Our experiments demonstrate that \ours{} achieves substantial advancements in generalist navigation. Without task-specific fine-tuning, \ours{} attains state-of-the-art or competitive performance across diverse public benchmarks for a variety of embodiments. On VLN-CE RxR~\citep{anderson2020rxr}, \ours{} improves performance in multi-camera settings (from $56.3\%$ to $64.4\%$ SR) and in single-camera settings (from $51.8\%$ to $57.4\%$ SR) compared to prior baselines. On HM3D-OVON~\citep{yokoyama2024hm3d}, our method achieves $45.2\%$ SR in a zero-shot setting, outperforming the previous fine-tuned SOTA method ($43.6\%$ SR). Similarly strong results are observed across various benchmarks in object searching, tracking, and autonomous driving. We further validate \ours{} through real-world experiments on multiple robotic platforms, including humanoid robots, quadrupeds, drones, and wheeled robots. These results underscore its strong generalizability and highlight promising progress toward generalist navigation.

\section{Related Works}

\textbf{Large models for Navigation.} Integrating Large Models (LLMs and VLMs) into robotic navigation has shifted the field from traditional learning-based methods toward leveraging pre-trained knowledge for open-world understanding and strong generalization capability. One straightforward approach~\citep{zhou2023navgpt, shah2022lmnav, qiao2023march} is to use off-the-shelf LLMs in a zero-shot manner. These works emphasize interpretability via chain-of-thought mechanisms~\citep{pan2023langnav, long2023discuss, lin2025navcot} and structured reasoning frameworks~\citep{chen2024mapgpt, qiao2025open}. However, abstracting dense visual information into text results in sparse environmental observations and is limited to static environments.
Another pathway~\citep{cheng2024navila, zhang2024navid, zhang2024uni, wei2025streamvln, wang2025trackvla} involves end-to-end fine-tuning of video-based or image-based~\citep{zhou2025navgpt2, zheng2024towards, zhang-etal-2025-mapnav} vision-language models with navigation data to enable VLMs to master navigation capabilities. However, existing methods mostly focus on isomorphic embodiments, overlooking the potential training synergy between different embodiments and tasks. In this work, we make an initial effort to extend the navigation policy to a broader domain of cross-embodiment and cross-task navigation.

\textbf{Cross-embodiments Navigation.}
The development of navigation models that generalize across diverse embodiments—varying in shape, size, and sensor configurations—remains a significant challenge in embodied AI. Recent efforts \citep{shah2023gnm,shah2023vint,yang2024pushing,wang2020environment,eftekhar2024one,hirose2023exaug,putta2024embodiment,curtis2024embodiment,wang2025x,zhang2025nava} have demonstrated the potential of transformer-based policies trained on large-scale, cross-embodiment datasets to achieve robust performance across various robotic platforms without the need for manual alignment of observation and action spaces.
% However, there are two main challenges in building a cross-embodiment navigation model. The first lies in the difference in action space. Prior works in vision-language navigation often discretize actions into a navigation graph or predefined choices such as moving forward by a fixed distance or turning by a fixed angle, which are not suitable for UAVs and autonomous driving. The second challenge is the variability in sensor configurations: different embodiments may have a different number of cameras placed in different orientations, making it harder for the model to integrate observations from multiple views in all embodiments. 
However, these models often process multimodal inputs without incorporating explicit spatial and temporal cues, which can lead to ambiguities due to the differing geometric interpretations of data from various embodiments. This approach may also result in data inefficiencies and limited generalization to out-of-distribution embodiments. (\cite{eftekhar2024one,wang2025rethinking})
In contrast, \ours % unifies the action space into trajectory prediction for all embodiments and
introduces temporal-spatial indicator tokens that encode the observation configurations, enabling the model to better interpret multimodal inputs across different embodiments. 
% Through extensive experiments on vision-and-language navigation and autonomous driving tasks, we demonstrate that our method achieves state-of-the-art performance and exhibits strong generalizability to diversified embodiments.

\textbf{Cross-task Navigation.}
Recent advances in embodied AI \citep{o2024open, team2024octo, kim2024openvla, bjorck2025gr00t, black2024pi_0, intelligence2025pi_, bu2025univla, bu2025agibot, qu2025embodiedonevision} demonstrate that generalist models built upon foundation models can effectively transfer knowledge across diverse tasks. In the domain of navigation, prior studies \citep{zhou2024same, wang2022towards, long2024instructnav, song2025towards, zhang2024uni, gao2025octonav, yin2025unigoal, ruan2025reactive} have shown that integrating data from different categories of navigation tasks can lead to stronger performance across various navigation scenarios. An early work, VIENNA \citep{wang2022towards}, leveraged reinforcement learning to train an agent in simulators. More recently, Uni-Navid \citep{zhang2024uni} developed a generalist model based on video-based vision-language models (VLMs) that learns generalized navigation skills through cross-task learning across four types of tasks: vision-and-language navigation, object-goal navigation, embodied question answering, and human following. However, these approaches are limited to constrained settings (\textit{e.g.}, indoor controllable environments), whereas our work extends to broader scenarios (\textit{e.g.}, including autonomous driving and UAV navigation) and unifies all tasks under a general formulation. In this framework, the model takes RGB videos and natural language instructions as input and outputs an executable trajectory.

% OctoNav (\cite{gao2025octonav}) develops a unified benchmark that encompasses object goal navigation, point goal navigation, vision-language navigation, image goal navigation, and instance-image goal navigation, and proposes a VLA-based model to accomplish these tasks. 
% \section{Task: Cross-Embodiment and Cross-Task Navigation}

\begin{figure}[t]
  \centering
    \includegraphics[width=\linewidth]{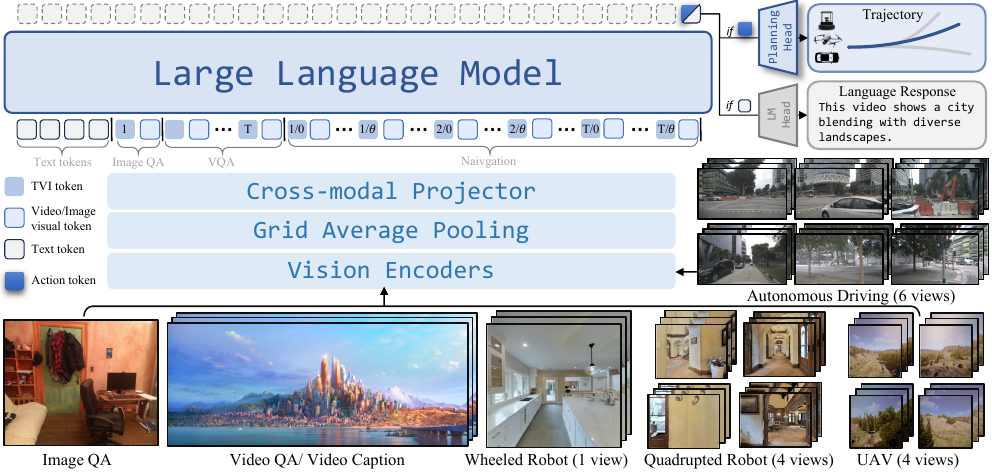}
  \caption{\textbf{Pipeline of \ours{}.} Our method provides a unified framework for handling multiple tasks, including Image QA, Video QA, and Navigation. We organize text tokens and visual tokens using temporal-viewpoint indicator tokens (sec.~\ref{sec:TVI_Tokens}), as described in Section~\ref{sec:llm_forwarding}. For question answering, our model employs a conventional language modeling head in an autoregressive manner, while for navigation, it uses a planning head to directly predict trajectories.}
  \label{fig:pipeline}
\end{figure}

\section{Method}

\textbf{Generalist Navigation Task}. We consider a general navigation setting in which a mobile embodiment is given a textual instruction $L$ and a sequence of images $I_{1:T}^{1:N} \in \mathbb{R}^{W \times H \times 3}$, captured on-the-fly from $N$ different cameras at time steps $\{1,...,T\}$. Given these observations and the instruction, our model $\pi$ is required to predict a navigation trajectory $\tau = \{\mathbf{a}_1, \mathbf{a}_2,...\}$, where each $\mathbf{a} \in \mathbb{R}^4 = (x, y, z, \theta)$ represents a position and orientation waypoint. Note that $z$ is only used when the embodiment is a UAV, and $\theta$ denotes the yaw angle (since our task does not require agile flight motions, the yaw angle suffices). The model drives the mobile embodiment to fulfill the instruction according to the mapping $\pi(L, I_{1:T}^{1:N}) \mapsto \tau_T$.

% \in \mathbb{R}^{P \times C}$ ($P$ is set to 576) and $C$ is the embedding dimension)
\textbf{Basic Architecture.} We extend vanilla video-based vision-language models (VLMs) \citep{li2023llama,shen2024longvu} to a dual-branch architecture for both navigation and question-answering \citep{wang2025trackvla}. For navigation, we first encode the observed images $I_{1:T}^{1:N}$ using vision encoders and a cross-modality projector~\citep{liu2023llava} to obtain visual tokens $E_{1:T}^{1:N}$. The instruction is embedded following common practices in existing language models~\citep{liu2023llava} to produce language tokens $E_L$. The visual tokens are then organized via temporal-viewpoint indicator tokens (sec.~\ref{sec:TVI_Tokens}) and budget-aware temporal sampling (sec.~\ref{sec:BATS}), concatenated with the language tokens, and fed into a large language model to predict the action token. This token is subsequently decoded by a planning model to generate a waypoint-based trajectory.
\begin{equation}
\begin{split}
    % E_{1:T}^{1:N} &= \text{Encoder}(I_{1:T}^{1:N}), \\
    E_T^A &= \text{LLM}({E_{1:T}^{1:N}, E_L}), \\
    \tau_T &= \text{ActionModel}(E_T^A).
\end{split}
\end{equation}
For the question-answering task, we follow existing methods~\cite{liu2023llava} and predict the next token in an auto-regressive manner. As in existing works~\citep{zhang2024navid,zhang2024uni,wang2025trackvla,cheng2024navila}, our model enables the co-tuning of both navigation and QA samples.

\subsection{Navigation Foundation Model}
\label{sec:nfm}

% It has been studied~\citep{kim2024openvla,tong2024eyes} that using both DINOv2 and SigLIP leads to significant performance improvements in vision-centric tasks, as DINOv2 captures fine-grained geometric features while SigLIP provides semantic understanding.

\textbf{Observation Encoding.}
Given captured egocentric RGB sequences $I_{1:T}^{1:N} \in \mathbb{R}^{W \times H \times 3}$ from $N$ multi-camera views at time step $T$, we employ pre-trained visual encoders (DINOv2~\citep{oquab2023dinov2} and SigLIP~\citep{zhai2023siglip}, a widely used recipe~\citep{kim2024openvla,tong2024eyes}) to extract visual features $\mathbf{V}_{1:T}^\text{dino/SigLIP} \in \mathbb{R}^{P \times C}$, where $P$ is the number of patches (set to 576) and $C$ represents the embedding dimension. For token savings and computational efficiency, we directly concatenate $V_{1:T}^\text{dino}$ and $V_{1:T}^\text{siglip}$ along the channel dimension and denote the resulting representation as $V_{1:T}$.
During navigation, on-the-fly captured videos leads an extensive number of frames, which subsequently produce an extensive set of visual features. To address this, we employ a grid pooling strategy~\citep{zhang2024navid,zhang2024uni} (Figure ~\ref{fig:pipeline}, Grid Average Pooling) on the visual features to generate more compact representations. Specifically, we utilize two resolution scales:
\begin{flalign}
\mathbf{V}^{\textcolor{blue}{\text{fine}} / \textcolor{mygreen}{\text{coarse}}} &= \text{GridPool}(\mathbf{V}, \textcolor{blue}{\frac{64}{P}} or \textcolor{mygreen}{\frac{4}{P}}),
\end{flalign} 
where $V^\text{fine}\in\mathbb{R}^{64 \times C}$ provides fine-grained observations, while $V^\text{coarse}\in\mathbb{R}^{4 \times C}$ offers coarse-grained observations. In this case, we use fine-grained features $V_\text{fine}$ for the latest navigation observation and image QA (at time step $T$), while using coarse-grained features for navigation history and video data (across time steps ${1:T}$).
Finally, following established VLMs~\citep{liu2023llava,li2023llama}, we use a cross-modality projector $\mathcal{P}(\cdot)$ (a 2-layer MLP) to project visual features into the latent space of the Large Language Model: $\mathbf{E}^V_T = \mathcal{P}(V_{1:T}^{1:N})$.

% To optimally balance token length and performance, we empirically use fine-grained features $V_\text{fine}\in\mathbb{R}^{64 \times C}$ for the latest tracking observation to enhance target identification, while coarse-grained tokens are used for historical tracking and VQA-based recognition.

% \begin{flalign}
% \mathbf{V}^{\text{fine}/\text{coarse}} &= GridPool(\mathbf{V}, \frac{64}{N}) \\
% \mathbf{V}^\text{coarse} &= GridPool(\mathbf{V}, \frac{4}{N})
% \end{flalign}

% To ensure consistent inference speed during tracking, we employ a sliding window mechanism to retain only the latest $k$ frames (set to 32 in our implementation). For embodied visual tracking, we structure the visual token sequence as: $\mathcal{V}^\text{track}_T = \{\mathbf{V}^\text{coarse}_{T-k},..., \mathbf{V}^\text{coarse}_{T-1}, \mathbf{V}^\text{fine}_{T}\}$, while for the video question answering (VQA) recognition task, we construct the sequence as: $\mathcal{V}^\text{VQA}_T = \{\mathbf{V}^\text{coarse}_{1},..., \mathbf{V}^\text{coarse}_T\}$. 

\begin{wrapfigure}{r}{0.4\linewidth}
  \centering
    \includegraphics[width=\linewidth]{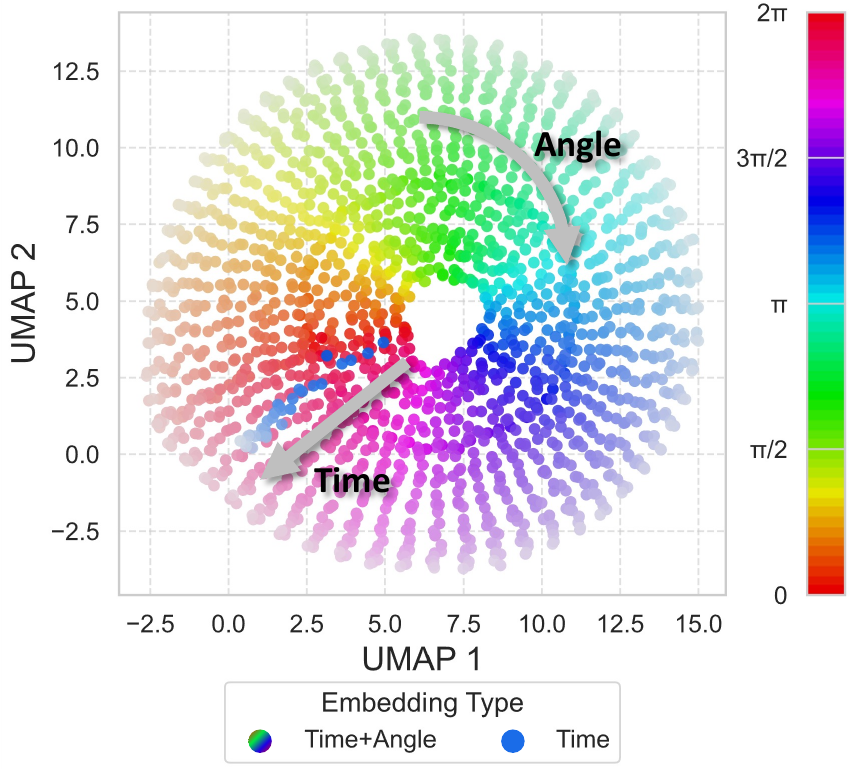}

  \vspace{-3mm}
  \caption{\textbf{Visualization of Temporal-Viewpoint Indicator (TVI) tokens.} We employ a clustering algorithm~\citep{mcinnes2018umap} to map high-dimensional embeddings into a 2D space.}
  \label{fig:tvi_tokens}
  \vspace{-6mm}
\end{wrapfigure}

% \vspace{-10mm}

\subsubsection{Temporal-Viewpoint Indicator (TVI) Tokens.} 
\label{sec:TVI_Tokens}
Given that visual tokens do not inherently incorporate viewpoint and temporal information, a key challenge in multi-view navigation models lies in enabling the LLM to discern which tokens correspond to different timesteps or distinct camera viewpoints. Previous approaches were limited to either specific camera configurations or embodiments \citep{long2024instructnav,gao2025octonav} or simply concatenated tokens from all viewpoint images \citep{zheng2024towards,fu2025orion}, thereby overlooking the flexibility of LLM token organization. To enable flexible processing of arbitrary camera arrangements, we introduce temporal-viewpoint indicator tokens, inspired by the demonstrated effectiveness of specially designed tokens for time/modality/task identification \citep{guo2025seed1,chen2023minigpt}, an approach that has been widely recognized to facilitate LLM learning. In our setting, the indicator tokens are used in diverse tasks, including image QA, video QA, and navigation, which should meet three important attributes:

\begin{itemize}
    \item \textbf{\textit{Viewpoint-Awareness}}: The token's angle embedding must preserve the circular continuity of azimuthal angles (\textit{e.g.}, $0\equiv2\pi$), ensuring that the distance metric between embeddings reflects geometric proximity (\textit{e.g.}, $d(0,\epsilon) < d(0,\pi)$ when $\epsilon\ne\pi$).
    \item \textbf{\textit{Time-Awareness}}: The token must uniquely identify the temporal order of frames across all camera views, while maintaining robustness to irregular sampling intervals.
    \item \textbf{\textit{Separability}}: The indicator tokens may encode either viewpoint or temporal information (for video QA) or may exclude such information entirely (for image QA).
\end{itemize}

To meet these requirements, our Temporal-Viewpoint Indicator (TVI) tokens $\mathbf{E}_\text{TVI} \in \mathbb{R}^C$ (where timestep and view angle are denoted as $t$ and $\phi$, respectively) consist of three types of embeddings: angle embedding $\text{AnglePE}(\phi)\in \mathbb{R}^C$, time embedding $\text{TimePE}(t) \in \mathbb{R}^C$, and a learnable base embedding $\mathbf{E}_\text{Base} \in \mathbb{R}^C$:
\begin{equation}
    \label{equ:TVI_Tokes}
    \mathbf{E}_\text{TVIT} = 
\begin{cases}
    \mathbf{E}_\text{Base} +  \mathcal{P}_\text{time}(\text{TimePE}(t)) + \mathcal{P}_\text{angle}(\text{AnglePE}(\phi)), & \text{if Navigation} \\
    \mathbf{E}_\text{Base} + \mathcal{P}_\text{time}(\text{TimePE}(t)),  & \text{if Video QA}\\
    \mathbf{E}_\text{Base}, & \text{if Image QA}
\end{cases}
\end{equation}
where $\text{AnglePE}(\phi)$ is implemented using a concatnation of sinusoidal position encodings~\citep{vaswani2017attention} applied to the cosine and sine values of the azimuthal angles separately, and $\text{TimePE}(t)$ is implemented as a sinusoidal position encoding of $t$. Here, $\mathcal{P}_\text{time}$ and $\mathcal{P}_\text{angle}$ are both implemented as two-layer MLPs (similar in design to those used in~\cite{liu2023llava}).
For different tasks and TVI tokens, we employ different combinations of indicator token components to represent the attributes of various visual tokens. For the navigation task, we include both temporal and viewpoint information. For the video QA task, we incorporate temporal information. For the image QA task, we use only $E_\text{Base}$ as an indicator that the subsequent tokens are visual tokens. This strategy offers a flexible approach to organizing significantly different sample types and facilitates LLM learning (Sec.~\ref{sec:llm_forwarding}). We provide a plot of the clustering results~\citep{mcinnes2018umap} of TVI Tokens in Figure~\ref{fig:tvi_tokens}, where we observe that the tokens are distinguished from one another according to the viewpoint $\theta$ (represented by a rainbow colorbar) and the timestep $t$ (represented by color value).

Note that while alternative techniques for inovling multi-view infomration such as positional encoding exist~\citep{zheng2024towards}, we empirically find (Table~\ref{tab:ablation_study_BATS_TVI}) that leveraging additional indicator tokens demonstrates the most robust performance during training and strong performance in evaluation. We believe this is because adding such tokens does not disrupt the off-the-shelf visual token space. This finding has also been reported in related literature~\citep{guo2025seed1,chen2023minigpt}.

\subsubsection{Budget-Aware Temporal Sampling (BATS). } 
\label{sec:BATS}
During navigation, on-the-fly captured video can generate an excessive number of visual tokens, increasing both inference and training time and hindering real-world deployment. Previous methods address this challenge in two ways: (1) Token Merging \citep{zhang2024uni}, which introduces additional computational overhead during training and leads to inconsistent inference speeds during evaluation; (2) Uniform Sampling \citep{cheng2024navila}, which often fails to adequately capture recent observations due to a lack of short-term context. Moreover, in scenarios involving variable camera-view settings (where the number of frames increases significantly) both strategies require additional modifications.
\begin{figure}[t]
  \centering
    \includegraphics[width=\linewidth]{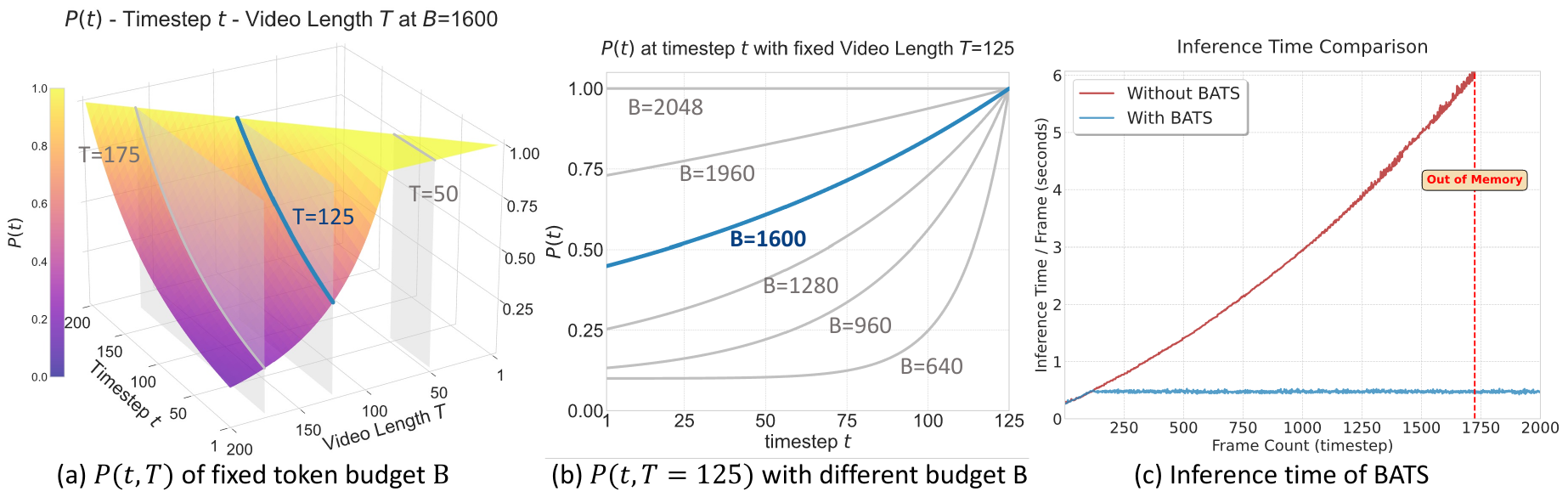}
  \caption{\textbf{Visualization of BATS and corresponding time cost.} (a) Given a fixed token budget $B = 1600$, we illustrate the sampling probability at different timesteps $t$ for the latest timestep $T$. (b) Given a maximum timestep $T = 125$, we plot the sampling probability across different timesteps $t$ under varying token budgets $B$. (c) We compare the inference time when using BATS versus not using BATS (keeping all frames).}
  \label{fig:BATS}
\end{figure}

To this end, we propose \emph{Budget-Aware Temporal Sampling (BATS)}, which is designed for (a) practical purposes (i.e., constraining the maximum token length to accommodate inference speed and GPU memory limitations), (b) retaining more recent information to enhance understanding and planning while preserving sufficient historical context for navigation, and (c) direct adaptability to varying numbers of cameras. Specifically, given a token budget $B_\text{token}$ and a multi-view video sequence ${I_{1:T}^{1:N} \in \mathbb{R}^{W \times H \times 3}}$, we employ an exponential growth based sampling probability $P(t)$, which is inspired by the “forgetting curve”. In this case, when the number of captured frame tokens exceeds the token budget, we compute a sampling probability for each frame:
\begin{equation}
    \label{equ: forgetting curve}
    % \begin{split}
        P(t) = (1-\epsilon)e^{k(t-T)/T} + \epsilon,  \quad k > 0,
    % \end{split}
\end{equation}
where the $\epsilon$ (we use $\epsilon = 0.1$) ensures that the lower bound of sampling probability is in the approximate range and the $k$ denotes the exponential decay rate. Therefore the expected number of sampled frames can be computed as:
\begin{equation}
    \label{eq:bats}
   \mathbb{E}_\text{frames}  \approx \int_{0}^{T}P(t) dt = (1-\epsilon) \frac{1 - e^{-k}}{k} T + \epsilon T   \\
\end{equation}

We constrain the expected number of tokens $((\textcolor{mygreen}{4}+1)\mathbb{E}_\text{frame}+(\textcolor{blue}{64}+1))N$ to be no larger than $B_\text{token}$. This implies $\mathbb{E}_\text{frame}\le\frac{B_\text{token}-(\textcolor{blue}{64}+1)N}{(\textcolor{mygreen}{4}+1)N}$, and with sufficiently large number of frames $T$, the number of sampled frames will converge to the expectation (Figure~\ref{fig:BATS} (c)).  
We can offline calculate $k$ for different $T$ using Brent’s method~\citep{brent2013algorithms}, leading correspongding $P(t)$ (Equation~\ref{equ: forgetting curve}). Note that since we set the lower-bound probability $\epsilon$, Equation~\ref{eq:bats} may become unsolvable for very large $T$  
% $T < B_{\text{frame}}/\epsilon$ 
(\textit{e.g.}, $T=1120$ under a four-camera setup with a token budget $B_\text{token} = 2048$). However, this situation rarely occurs (for the list task in Figure~\ref{fig:teaser_comp}), as most timesteps are approximately $122$ steps in VLN-CE RxR~\citep{anderson2020rxr}. We provide the details of using BATS in Appendix~\ref{appendix:BATS}.

We plot the distribution of timestep sampling probability and time efficiency in Figure~\ref{fig:BATS}. It can be observed that our method smoothly obtains reasonable $P(t)$ under different token budgets $B$ and timesteps $T$. With a higher token budget, the BATS strategy adaptively samples more historical tokens; even with fewer tokens, our strategy still maintains a reasonable lower bound. Furthermore, we note that BATS maintains a stable inference speed throughout the entire navigation process.

\begin{figure}[t]
  \centering
    \includegraphics[width=0.7\linewidth]{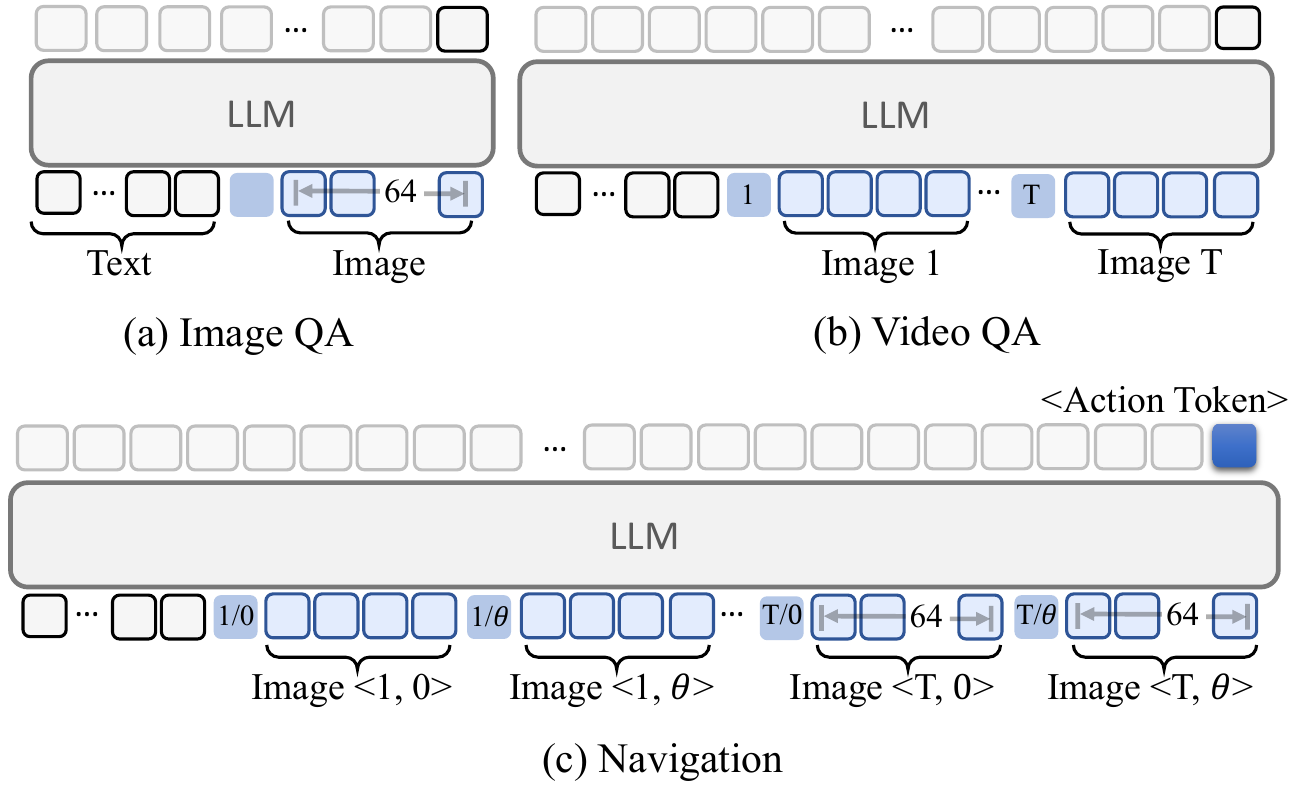}
  \caption{\textbf{Token Organization Strategy of \ours Across Different Tasks.} (a) For image question answering, fine-grained visual tokens are utilized, incorporating only the base embedding of TVI tokens. (b) For video question answering, coarse-grained visual tokens are employed, which include both the base embedding and the time embedding of TVI tokens. (c) For navigation, both coarse-grained and fine-grained visual tokens are used, integrating the base, time, and angle embeddings of TVI tokens.}
  \label{fig:token_organ}
\end{figure}

\subsubsection{LLM Fowarding} 
\label{sec:llm_forwarding}
\textbf{Token Organization.} After obtaining the visual tokens $E_{1:T}^{1:N}$(sampled via BATS, Sec.~\ref{sec:BATS}) and the language tokens $E_L$, we organize these tokens using TVI Tokens (Sec.~\ref{sec:TVI_Tokens}) for forwarding through the LLM. We provide a detailed illustration of the token organization strategy for different tasks in Figure~\ref{fig:token_organ}. For Image QA, we use $E_\text{Base}$  along with fine-grained visual tokens (64 tokens per image) to represent the image. For Video QA, we incorporate $\mathbf{E}_\text{Base} + \mathcal{P}_\text{time}(\text{TimePE}(t))$
 to encode temporal information for each frame, and employ coarse-grained visual tokens (4 tokens per frame) to avoid an excessive number of tokens. For navigation, we use $\mathbf{E}_\text{Base} +  \mathcal{P}_\text{time}(\text{TimePE}(t)) + \mathcal{P}_\text{angle}(\text{AnglePE}(\phi))$ to represent both temporal and viewpoint information. Here, fine-grained visual tokens are used for the most recent observations, while coarse-grained tokens are utilized for historical observations. Our token organization strategy enhances the LLM's understanding of the input tokens and supports a unified framework for Image QA, Video QA, and navigation tasks.

\textbf{Trajectory prediction.} For the navigation task, given the predicted action hidden state  $E^\text{A}_T$ from the forward pass of the LLM, we apply a plannning model $\mathcal{A}_\theta$ (implemented as a three-layer MLP) to extract the trajectory information ${\tau}_T$. Note that the original trajectory may range from a few meters (indoor navigation) to tens of meters (autonomous driving and drones). In this case, directly predicting the raw trajectory could lead to divergence in the waypoint distribution. Therefore, following previous methods~\cite{shah2023gnm}, we normalize the waypoints of trajectories to a distribution of $[-1, 1]$ using a task-specific scaling factor $\alpha_\text{task}$. Here, we use three different scaling factors for indoor navigation, UAVs, and cars, as shown in Appendix~\ref{appendix:action}. We can formulate the trajectory prediction as follows:
\begin{equation}
\label{eq:action_model}
\tau_T = \{\mathbf{a}_1,...,\mathbf{a}_M\}_T = \alpha_\text{task} \cdot \mathcal{A}_\theta(E^\text{A}_T),
\end{equation}
where $M$ is set to $8$, and the normalized trajectory is rescaled to absolute values by multiplying by $\alpha_\text{task}$. The trajectory loss is computed using the mean squared error (MSE) $L_\text{nav} = \text{MSE}(\tau^\text{idx},\tau^\text{idx}_\text{gt})$, there idx denotes the valid action indices. For wheeled robots/car embodiments, $\mathbf{a}^{\text{idx}} = (x, y, \theta)$; for UAVs, $\mathbf{a}^{\text{idx}} = (x, y, z, \theta)$. For the question-answering task, we employ the cross-entropy loss $L_\text{QA}$ under a next-token-prediction supervision framework. Given a batch containing both navigation and QA samples, the total loss is defined as $L = \beta L_\text{nav} + L_\text{QA}$. Here, $\beta$ is a constant scaling factor (set to 10) used to amplify the navigation loss, which tends to be numerically small since it is derived from mean squared error.

\subsection{Data}

\begin{wrapfigure}{r}{0.25\linewidth}
  \centering
    \includegraphics[width=\linewidth]{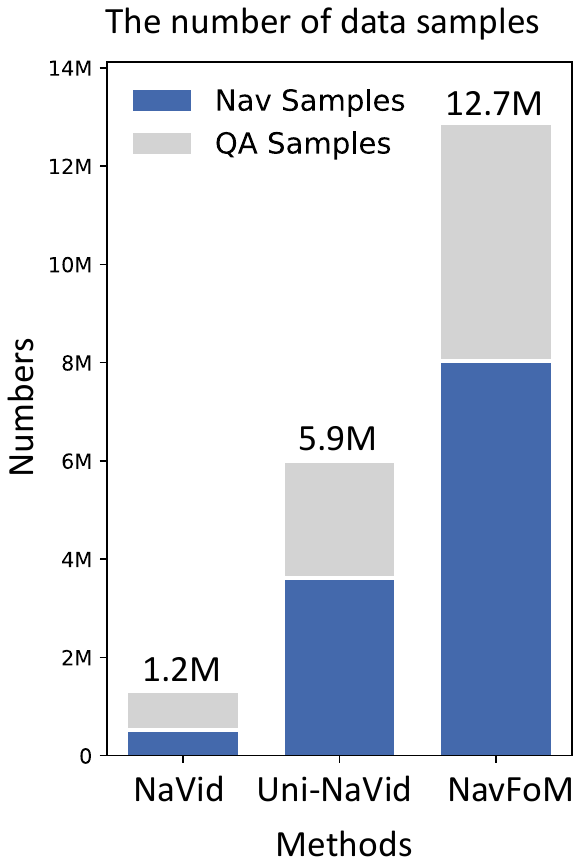}
  \caption{Comprasion of number of training samples with previouse methods.}
  \label{fig:data_comp}
\end{wrapfigure}

% \begin{wrapfigure}{r}{0.45\linewidth}
%     \centering
%     \vspace{-10pt} 
%     \includegraphics[width=\linewidth]{figure/action_model.pdf}
%     % \vspace{-20pt}
%     \caption{Anchor-based Diffusion Action Model.}
%     \vspace{-1em}
%     \label{fig:action_model}
% \end{wrapfigure}

To fine-tune \ours, we collect and process a large set of comprehensive and diverse training samples, totaling 12.7 million instances. These include 8.02 million navigation samples, 3.15 million image-based question-answering samples, and 1.61 million video-based question-answering samples. Our training samples are large than previouse methods~\citep{zhang2024navid,zhang2024uni}. The navigation samples are collected and processed across diverse mobile embodiments (wheeled robots, quadruped robots, UAVs, and cars) covering a variety of tasks including vision-and-language navigation, object-goal navigation, active visual tracking, and autonomous driving. All navigation data are collected in a unified manner, including previously captured videos (from both single and multiple cameras), instructions, and predicted trajectory waypoints.
As for the question-answering samples, we gather image-based QA and video-based QA data from off-the-shelf datasets following existing video-based Vision-Language Models (VLMs)~\citep{shen2024longvu, li2023llama}.
Further details regarding the navigation samples are elaborated below:

\textbf{Vision-and-Language Navigation} (3.37 M) requires an agent to interpret natural language instructions and egocentric visual observations, align the instructions with visual inputs, and plan subsequent actions to reach described landmarks. Following a broad definition of VLN~\citep{zheng2024towards, wang2025rethinking, zhou2024same}, we consider both indoor environments (e.g., VLN-CE on R2R~\citep{krantz2020beyond} and RxR~\citep{anderson2020rxr}) and outdoor environments (e.g., OpenUAV~\citep{wang2024openuav}), deployed on robots and unmanned aerial vehicles (UAVs), respectively.
For VLN-CE on R2R and RxR (2.94 M), we capture multi-view RGB videos, annotated instructions, and trajectory data while the robot follows the ground-truth path. The multi-view RGB setup consists of a fixed front-view camera and randomly sampled surrounding cameras (ranging from one to eight). Camera heights are randomized between 0.6 m and 1.5 m, and the horizontal fields of view (HFoV) vary between 75$^\circ$ and 120$^\circ$. 
For the OpenUAV dataset (429 K), we record camera streams from the front, left, right, and rear views for all sequences. Other randomization strategies remain consistent with those used in the VLN-CE tasks.

% randomly sample camera views surrond the robots , 

% we process XM navigation samples of VLN datasets, VLN-CE R2R~\cite{Krantz2020BeyondTN} and RxR~\cite{ku2020room}, that focus on continuous environments. }

\textbf{Object Goal Navigation} (1.02 M) requires a robot to explore an unseen environment and identify a described target. For the object goal navigation dataset, we follow the method of~\citep{zhang2024uni} by collecting successful episodes from L3MVN~\citep{yu2023l3mvn}, a heuristic-designed approach that explicitly models the exploration and identification stages. Our data are collected from HM3D ObjectNav~\citep{habitat19iccv} episodes, which require the agent to locate objects from a predefined category set (e.g., \textit{sofa}, \textit{chair}, and \textit{bed}). Nevertheless, experiments show that our method generalizes to state-of-the-art open-vocabulary object goal searching, as presented in Table~\ref{tab:openvocab-objnav}. Note that we do not employ multiple cameras or camera randomization, as we aim to maintain the same visual observation configuration as L3MVN.

\textbf{Active Visual Tracking} (897K)~\citep{islam2019person, francis2023principles, wang2025trackvla} requires the robot to distinguish the target within dynamic and crowded environments. The target is specified via textual instructions, e.g., “Follow the man in the blue t-shirt.” The agent must recognize the appearance of the human, follow the correct person according to the instructions, and maintain an appropriate distance while avoiding obstacles. For this task, we use data from EVT-Bench, consistent with~\citep{wang2025trackvla}, which involves diverse indoor environments and hundreds of avatars with corresponding descriptions. We also incorporate camera randomization, as described in our VLN data collection process.

% track and follow a human target with a specific description in dynamic and crowded environments, \egno, \textit{``Follow the man in the blue t-shirt.''}. 

\textbf{Autonoums Driving} (681K)~\citep{related_work_driving_UniAD, related_work_driving_DiffusionDrive} requires an agent to generate a safe, comfortable, and kinematically feasible trajectory for navigating complex and dynamic real-world environments. This task evaluates the agent's ability to continuously perceive its surroundings, anticipate the future movements of other traffic participants, and make robust sequential decisions to avoid collisions while progressing toward a destination. Here, we process 27K and 654K samples sourced from nuScenes~\citep{Driving_Dataset_Nuscenes} and OpenScene~\citep{openscene2023}, respectively. We directly record the original multi-view images, instructions, and vehicle state information from the dataset. Note that we do not collect explicit surrounding information (such as lane details), in contrast to common autonomous driving baselines~\citep{chen2024drivinggpt, LAW}.

% and ImpromptuVLA~\cite{chi2025impromptu} datasets. This collection encompasses a variety of camera configurations (1, 3, 5, 6, 7, and 8 views) and provides comprehensive coverage of driving environments, including rich real-world scenarios, challenging situations that demand dynamic decision-making, and unstructured corner cases such as unclear road boundaries, temporary traffic rules, unconventional obstacles, and adverse road conditions.

\textbf{Web-Video Navigation.} (2.03M) We also leverage the Sekai dataset~\citep{li2025sekai}, which provides a collection of approximately 182K YouTube videos with corresponding instructions (generated by VLMs~\citep{bai2025qwen2}) and trajectories (generated by SLAM systems~\citep{li2025megasam}). Although these navigation samples contain imperfect instructions and trajectories, they remain valuable for incorporating real-world navigation scenarios. Similar findings have been reported in~\citep{cheng2024navila, wei2025streamvln}.

% web videos and corresponding trajectories from dataset~\addcite{}, which include navigation videos along with corresponding instructions (generated by VLMs~\addcite{}) and trajectories (generated by SLAM systems~\addcite{}). Although these samples contain noise (imperfect trajectories and instructions), they are valuable for incorporating significant diversity across different scenes with real captured images.

% 3.15M IQA and 1.61M VQA

\textbf{Open-World Question-Answering.} (4.76M) Following existing video-based VLMs~\citep{li2023llama, shen2024longvu, wang2025trackvla}, we collect 3.15M image QA samples and 1.61M video QA samples, which encompass rich and comprehensive knowledge for open-world understanding.

% We follow previsous scuccess of RT-2~\addcite{}, Uni-NaVid~\addcite{} and Pi-0.5~\addcite{}, we co-tune open-world QA samples with navigation samples to maintat the open-world knowelage of \ours{}.

\subsection{Implementation Details} 

% \textbf{Architecture Detials.}

\textbf{Training Configurations.} Our model is trained on a cluster server equipped with 56 NVIDIA H100 GPUs for approximately 72 hours, resulting in a total of 4,032 GPU hours. For question-answering data, all frames are sampled at 1 FPS to reduce redundancy between consecutive frames. For discrete navigation data (e.g., Habitat environments~\cite{habitat19iccv}), we sample each step after the robot performs a discrete action (See Appendix~\ref{appendix:action} for details on how discrete actions are modified into trajectories.). 
For continuous navigation environments (e.g., EVT-Bench~\cite{wang2025trackvla}, autonomous driving~\citep{caesar2020nuscenes, openscene2023}), data are sampled at 2 FPS to avoid redundancy. During training, the vision encoders (DINOv2~\cite{oquab2023dinov2} and SigLIP~\cite{zhai2023siglip}) and the large language model (Qwen2-7B~\cite{qwen2}) are initialized with their default pre-trained weights. Following the training paradigm of VLM~\citep{liu2023llava}, we fine-tune only the designated trainable parameters for a single epoch.

\clearpage

\begin{wrapfigure}{r}{0.5\linewidth}
  \centering
    \includegraphics[width=\linewidth]{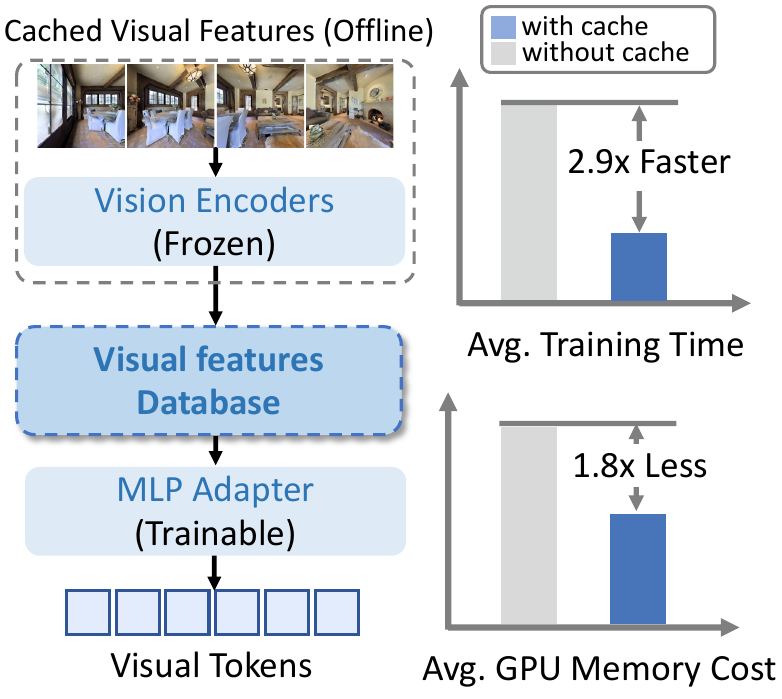}
  \caption{\textbf{Offline Visual Feature Cached.} We pre-computed video frames and navigation hisitroy and saved as corase visual tokens.}
  \label{fig:token_cache}
\end{wrapfigure}

% \begin{wrapfigure}{r}{0.45\linewidth}
%     \centering
%     \vspace{-10pt} 
%     \includegraphics[width=\linewidth]{figure/action_model.pdf}
%     % \vspace{-20pt}
%     \caption{Anchor-based Diffusion Action Model.}
%     \vspace{-1em}
%     \label{fig:action_model}
% \end{wrapfigure}

\textbf{Accelerating Training by Caching Visual Features.} Due to the long horizon of videos (hundreds of frames), encoding all images online in a large batch can be computationally expensive. To mitigate this issue, we leverage a visual feature caching mechanism~\citep{yan2022videococa} and construct a visual feature database (See Figure~\ref{fig:token_cache}). Note that we only cache coarse-grained visual tokens (4 tokens per frame), which require significantly less disk space compared to storing full videos, as a single episode of navigation typically produces dozens of videos. For image QA and the latest observation in navigation, we still use visual encoders online to extract fine-grained visual tokens (64 tokens per frame). This approach reduces training time (2.9× faster) and GPU memory usage (1.8× less).

% \textbf{Inference Detials.}
\section{Experiments}
\label{sec:exp}

\subsection{Experiment Setup}

To evaluate the performance of \ours{}, we conduct extensive experiments and ablation studies addressing three key aspects: (1) How does \ours{} perform on diverse navigation tasks across different benchmarks? (2) How well does \ours{} perform in real-world environments? (3) Are the key design components of our method effective? Our method is compared against strong baselines on each benchmark.

% Additional details are provided in the supplementary material.

\textbf{Benchmarks}. We evaluate our method on various navigation tasks, including VLN, searching, tracking, and autonomous driving, which are across different embodiments (e.g., egocentric, four-camera, six-camera, and eight-camera configurations). In all benchmarks, \ours{} uses only online-captured egocentric video (some from multi-view sources) and an instruction as input to predict the trajectory for the robot to execute. Given the diversity of benchmarks spanning multiple environments and simulators, we meticulously verify scene splits to ensure no overlap exists between training and validation scenes across all benchmarks.

\begin{itemize} 
\item \textbf{\textit{Vision-and-Language Navigation}}: We evaluate our method on the VAL-Unseen splits of the VLN-CE R2R~\citep{krantz2020beyond} and RxR~\citep{anderson2020rxr} benchmarks, which require the robot to follow instructions in unseen indoor environments. We also evaluate our method on the Open-UAV benchmark~\citep{wang2024openuav}, which requires the UAV to follow instructions in unseen outdoor environments.
\item \textbf{\textit{Object goal navigation}}: We follow previous methods~\citep{zhang2024uni, zhu2025mtu} to evaluate the generalizability of object-goal navigation on the HM3D-OVON dataset~\citep{yokoyama2024hm3d}, an open-vocabulary object navigation benchmark, in a zero-shot manner.
\item \textbf{\textit{Active Visual Tracking}}: We evaluate our method on EVT-Bench~\citep{wang2025trackvla}, a challenging benchamrak requrie the robot to distinguish and follow target within crowded environments.
\item \textbf{\textit{Autonomous Driving}}: We evaluate our method on mainstream benchmarks, namely nuScenes~\citep{Driving_Dataset_Nuscenes} and NAVSIM~\citep{Driving_Dataset_NAVSIM}, for open-loop and pseudo-simulation evaluation.

\end{itemize}

\textbf{Metrics.} To evaluate navigation performance, we follow the standard evaluation metrics~\citep{anderson2018vision}, including success rate (SR), oracle success rate (OS), success weighted by path length (SPL), normalized Dynamic Time Warping (nDTW) and navigation error from goal (NE). Espeecially for tracking task, we leverage tracking rate (TR)~\citep{puig2023habitat}, which measure the ratio of success tracking steps over all steps. Note that the success criteria change among different navigation tasks, we therefore use the default success criteria of each benchmark. 
For autonomous driving evaluation, we report L2 distance and Collision Rate (CR) for open-loop planning~\citep{Driving_Dataset_Nuscenes}. For closed-loop evaluation on NAVSIM, we use the PDM score (PDMS)~\citep{Driving_Dataset_NAVSIM}, which is a weighted combination of several sub-metrics: no at-fault collisions (NC), drivable area compliance (DAC), time-to-collision (TTC), comfort (Comf.), and ego progress (EP). 

% For video understanding evaluation,  we employ widely used metrics following existing works~\cite{azuma2022scanqa, li2023llama}.

\textbf{Deployment on Simulators.} For each navigation task, we adhere to the default settings established in prior works~\citep{krantz2020beyond,savva2019habitat,das2018embodied,islam2019person}. Our method takes as input online-captured RGB videos from a variable number of cameras (each capturing one frame after a step is taken) along with a textual instruction, and outputs the next trajectory (Equation~\ref{eq:action_model}). Note that for certain benchmarks in Habitat-Lab continuous environments that use discrete actions (such as \texttt{FORWARD}, \texttt{LEFT}, \texttt{RIGHT}, and \texttt{STOP}), we replace these discrete actions with trajectory-based actions. Further details are provided in the Appendix.

\textbf{Deployments on Real-World Environemnts.} For real-world deployment, we employ a remote server equipped with an NVIDIA RTX 4090 GPU to run \ours. The system processes observations, along with text instructions, and sends action commands to a local robot. Our experiments involve quadruped robots, humanoids, drones, and wheeled robots for trajectory execution. \ours requires at most 0.5 seconds to generate an eight-waypoint trajectory under a 1600-token budget. During navigation, the robot asynchronously compresses and uploads the latest observations to the model while concurrently executing actions. For different robots, we utilize their off-the-shelf local planners to steer them along the predicted trajectory. Please refer to the appendix for a detailed description of the real-world robot setup.

\subsection{Benchmark Results}

\begin{table}[t]
    \small
    \centering
    \caption{\textbf{Comparison on VLN-CE in Single-View and Multi-View Settings.} Here, S.RGB and M.RGB denote single-view and multi-view configurations, respectively. The symbol $^{*}$ indicates methods that utilize the waypoint predictor from~\citep{hong2022bridging}. Our method achieves SOTA performance in both single-view and multi-view settings (without fine-tuning on specific camera setups) and does not require additional inputs such as depth or odometry.}
    % \vspace{-5pt}
    \setlength{\tabcolsep}{1.4pt}
    \scalebox{0.9}{
{\fontsize{8pt}{9pt}\selectfont
\begin{tabular}{lcccclcccclcccc}
\toprule
Method & \multicolumn{4}{c}{Observation} & & \multicolumn{4}{c}{R2R Val-Unseen} & & \multicolumn{4}{c}{RxR Val-Unseen} \\
\cmidrule(lr){2-5} \cmidrule(lr){7-10} \cmidrule(lr){12-15}
& S.RGB & M.RGB & Depth & Odo. & & NE $\downarrow$ & OS $\uparrow$ & SR $\uparrow$ & SPL $\uparrow$ & & NE $\downarrow$ & SR $\uparrow$ & SPL $\uparrow$ & nDTW $\uparrow$ \\
\midrule
AG-CMTP~\citep{chen2021topological} & & \checkmark  & \checkmark & \checkmark & & 7.90 & 39.0 & 23.0 & 19.0 & & - & - & - & - \\
R2R-CMTP~\citep{chen2021topological} & & \checkmark  & \checkmark & \checkmark & & 7.90 & 38.0 & 26.0 & 22.0 & & - & - & - & - \\
HPN+DN$^{*}$~\citep{krantz2021waypoint} &  & \checkmark & \checkmark & \checkmark & & 6.31 & 40.0 & 36.0 & 34.0 & & - & - & - & - \\
CMA$^{*}$~\citep{hong2022bridging} & & \checkmark & \checkmark & \checkmark & & 6.20 & 52.0 & 41.0 & 36.0 & & 8.76 & 26.5 & 22.1 & 47.0 \\
VLN$\circlearrowright$BERT$^{*}$~\citep{hong2022bridging} & & \checkmark & \checkmark & \checkmark & & 5.74 & 53.0 & 44.0 & 39.0 & & 8.98 & 27.0 & 22.6 & 46.7 \\
Sim2Sim$^{*}$~\citep{krantz2022sim} & & \checkmark  & \checkmark & \checkmark & & 6.07 & 52.0 & 43.0 & 36.0 & & - & - & - & - \\
AO-Planner~\citep{chen2024affordances} & & \checkmark & \checkmark & & & 5.55 & 59.0 & 47.0 & 33.0 & & 7.06 & 43.3 & 30.5 & 50.1 \\
GridMM$^{*}$~\citep{wang2023gridmm} & & \checkmark  & \checkmark & \checkmark & & 5.11 & 61.0 & 49.0 & 41.0 & & - & - & - & - \\
Ego$^{2}$-Map$^{*}$~\citep{hong2023learning} & & \checkmark  & \checkmark & \checkmark & & 5.54 & 56.0 & 47.0 & 41.0 & & - & - & - & - \\
DreamWalker$^{*}$~\citep{wang2023dreamwalker} & & \checkmark  & \checkmark & \checkmark & & 5.53 & 59.0 & 49.0 & 44.0 & & - & - & - & - \\
Reborn$^{*}$~\citep{an20221st} & & \checkmark  & \checkmark & \checkmark & & 5.40 & 57.0 & 50.0 & 46.0 & & 5.98 & 48.6 & 42.0 & 63.3 \\
ETPNav$^{*}$~\citep{an2024etpnav} & & \checkmark  & \checkmark & \checkmark & & 4.71 & 65.0 & 57.0 & 49.0 & & 5.64 & 54.7 & 44.8 & 61.9 \\
HNR$^{*}$~\citep{wang2024lookahead} & & \checkmark  & \checkmark & \checkmark & & \textbf{4.42} & 67.0 & 61.0 & 51.0 & & 5.50 & 56.3 & 46.7 & 63.5 \\
BEVBert$^{*}$~\citep{an2023bevbert} & & \checkmark  & \checkmark & \checkmark & & 4.57 & 67.0 & 59.0 & 50.0 & & - & - & - & - \\
HAMT+ScaleVLN$^{*}$~\citep{wang2023scaling} & & \checkmark  & \checkmark & \checkmark & & 4.80 & - & 55.0 & 51.0 & & - & - & - & - \\
\rowcolor{myblue}
\textbf{\ours (Four views)} & & \checkmark  &  &  & & 4.61 & \bf72.1 & \bf61.7 & \bf55.3 & & \bf4.74 & \bf64.4 & \bf56.2 & \bf65.8 \\
\midrule
LAW~\citep{raychaudhuri2021language} & \checkmark & & \checkmark & \checkmark & & 6.83 & 44.0 & 35.0 & 31.0 & & 10.90 & 8.0 & 8.0 & 38.0 \\
CM2~\citep{georgakis2022cross} & \checkmark & & \checkmark & \checkmark & & 7.02 & 41.0 & 34.0 & 27.0 & & - & - & - & - \\
WS-MGMap~\citep{chen2022weakly} & \checkmark & & \checkmark & \checkmark & & 6.28 & 47.0 & 38.0 & 34.0 & & - & - & - & - \\
% StreamVLN-Depth-Odo.~\citep{wei2025streamvln} & \checkmark  & & \checkmark & \checkmark & & \bf4.98 & 64.2 & \bf56.9 & 51.9 & & 6.22 & 52.9 & 46.0 & 61.9 \\
Seq2Seq~\citep{krantz2020beyond} & \checkmark & & \checkmark & & & 7.77 & 37.0 & 25.0 & 22.0 & & 12.10 & 13.9 & 11.9 & 30.8 \\
CMA~\citep{krantz2020beyond} & \checkmark & & \checkmark & & & 7.37 & 40.0 & 32.0 & 30.0 & & - & - & - & - \\
RGB-Seq2Seq~\citep{krantz2020beyond} & \checkmark & & & & & 10.10 & 8.0 & 0.0 & 0.0 & & - & - & - & - \\
RGB-CMA~\citep{krantz2020beyond} & \checkmark & & & & &  9.55 & 10.0 & 5.0 & 4.0 & & - & - & - & - \\
NaVid~\citep{zhang2024navid} & \checkmark & & & & &  5.72 & 49.2 & 41.9 & 36.5 & & 5.72 & 45.7 & 38.2 & - \\
Uni-NaVid~\citep{zhang2024uni} & \checkmark & & & & &  5.58 & 53.3 & 47.0 & 42.7 & & 6.24 & 48.7 & 40.9 & - \\
NaVILA~\citep{cheng2024navila} & \checkmark  & & & & & 5.22 & 62.5 & 54.0 & 49.0 & & 6.77 & 49.3 & 44.0 & 58.8 \\
StreamVLN-RGB-only~\citep{wei2025streamvln} & \checkmark  & &  &  & & 5.10 & 64.0 & 55.7 & 50.9 & & 6.16 & 51.8 & 45.0 & \bf62.1 \\
\rowcolor{myblue}
\textbf{\ours (Single view)} & \checkmark  & & & & & \bf5.01 & \bf64.9 & \bf56.2 & \bf51.2 & & \bf5.51 & \bf57.4 & \bf49.4 & 60.2 \\
\bottomrule
\end{tabular}}}
\vspace{-5pt}
\label{tab:r2r_rxr}
\end{table}

% anderson2020rxr
\textbf{VLN: Performence on VLN-CE}~\citep{krantz2020beyond, anderson2020rxr}. We begin by evaluating our method on the most widely used vision-and-language instruction benchmarks—VLN-CE R2R and VLN-CE RxR—with the results presented in Table~\ref{tab:r2r_rxr}. We report performance under both single-camera and four-camera settings (360$^\circ$ observations). Note that our model is not fine-tuned on any specific camera configuration; instead, visual tokens are directly organized using temporal-viewpoint indicator tokens (Figure~\ref{fig:token_organ}). Our method achieves state-of-the-art (SOTA) performance on both benchmarks across different camera settings. Under the most challenging condition—single-view VLN-CE RxR—our method improves success rate (SR) from $51.8\%$ to $57.4\%$. Notably, in multi-camera setups, our approach uses only four RGB cameras and attains an SR of $64.4\%$, outperforming previous SOTA methods ($56.3\%$ SR) that rely on RGB-D cameras and odometry information. This result clearly demonstrates the effectiveness of our method. We also observe a significant performance gain when transitioning from single-view to multi-view settings: an increase of $5.5\%$ on R2R-CE and $7.0\%$ on RxR-CE, respectively. This suggests that multi-view navigation foundation models represent a promising direction for future research. Furthermore, unlike other baselines, our method performs better on RxR than on R2R, even though RxR involves longer-horizon and more complex instructions. We attribute this to the more concrete and detailed descriptions in RxR instructions, which provide stronger contextual cues and help the model better distinguish target landmarks.

\begin{table}[t]
\centering
\caption{\textbf{Comprehensive results on \texttt{OpenUAV} benchmark with L1 level assistant.} \textbf{\texttt{Seen}} denotes the seen split, while \textbf{\texttt{UO}} and \textbf{\texttt{UM}} represent the Test Unseen Object Set and Test Unseen Map Set respectively. DA refers to a model trained using backtracking sampling-based data aggregation. The \textbf{best} and the \underline{second best} results are denoted by \textbf{bold} and \underline{underline}.}
\label{tab:openuav_combined_results}
\resizebox{\columnwidth}{!}{
\begin{tabular}{lc rrrr rrrr rrrr}
    \toprule
    \multirow{2}{*}{Method} & \multirow{2}{*}{Test Set} & \multicolumn{4}{c}{Full} & \multicolumn{4}{c}{Easy} & \multicolumn{4}{c}{Hard} \\
    \cmidrule(lr){3-6} \cmidrule(lr){7-10} \cmidrule(lr){11-14}
    & & NE$\downarrow$ & SR$\uparrow$ & OSR$\uparrow$ & SPL$\uparrow$ & NE$\downarrow$ & SR$\uparrow$ & OSR$\uparrow$ & SPL$\uparrow$ & NE$\downarrow$ & SR$\uparrow$ & OSR$\uparrow$ & SPL$\uparrow$ \\
    \midrule
    \multicolumn{14}{c}{\textit{OpenUAV Seen Set}} \\
    \midrule
    Human & Seen & 14.15 & 94.51 & 94.51 & 77.84 & 11.68 & 95.44 & 95.44 & 76.19 & 17.16 & 93.37 & 93.37 & 79.85 \\
    Random Action & Seen & 222.20 & 0.14 & 0.21 & 0.07 & 142.07 & 0.26 & 0.39 & 0.13 & 320.12 & 0.00 & 0.00 & 0.00 \\
    Fixed Action & Seen & 188.61 & 2.27 & 8.16 & 1.40 & 121.36 & 3.48 & 11.48 & 2.14 & 270.69 & 0.79 & 4.09 & 0.49 \\
    % \midrule
    CMA~\citep{wang2024openuav} & Seen & 135.73 & 8.37 & 18.72 & 7.90 & 84.89 & 11.48 & 24.52 & 10.68 & 197.77 & 4.57 & 11.65 & 4.51 \\
    TravelUAV~\citep{wang2024openuav} & Seen & 106.28 & 16.10 & 44.26 & 14.30 & 68.78 & 18.84 & 47.61 & 16.39 & 152.04 & 12.76 & 40.16 & 11.76\\
    TravelUAV-DA & Seen & \underline{98.66} & \underline{17.45} & \underline{48.87} & \underline{15.76} & \underline{66.40} & \underline{20.26} & \underline{51.23} & \underline{18.10} & \textbf{138.04} & \underline{14.02} & \textbf{45.98} & \underline{12.90} \\
    \rowcolor{myblue}
    \textbf{\ours (Four views)} & Seen & \textbf{93.05} & \textbf{29.17} & \textbf{49.24} & \textbf{25.03} & \textbf{58.98} & \textbf{32.91} & \textbf{53.16} & \textbf{27.87} & \underline{143.83} & \textbf{23.58} & \underline{43.40} & \textbf{20.80} \\
    \midrule
    \multicolumn{14}{c}{\textit{OpenUAV Unseen Set}} \\
    \midrule
    Random Action & UO & 260.14 & 0.16 & 0.16 & 0.16 & 174.10 & 0.48 & 0.48 & 0.48 & 302.96 & 0.00 & 0.00 & 0.00 \\
    Fixed Action & UO & 212.84 & 3.66 & 9.54 & 2.16 & 151.66 & 6.70 & 13.88 & 3.72 & 243.29 & 2.14 & 7.38 & 1.38 \\
    CMA~\citep{wang2024openuav} & UO & 155.79 & 9.06 & 16.06 & 8.68 & 102.92 & 14.83 & 22.49 & 13.90 & 182.09 & 6.19 & 12.86 & 6.08 \\
    TravelUAV~\citep{wang2024openuav} & UO & \underline{118.04} & \underline{22.42} & \underline{46.90} & \underline{20.51} & \underline{86.12} & \underline{24.40} & \underline{49.28} & \underline{22.03} & \underline{134.03} & \underline{21.43} & \underline{45.71} & \underline{19.75} \\
    \rowcolor{myblue}
    \textbf{\ours (Four views)} & UO & \textbf{108.04} & \textbf{29.83} & \textbf{47.99} & \textbf{27.20} & \textbf{70.51} & \textbf{32.54} & \textbf{50.72} & \textbf{29.54} & \textbf{133.01} & \textbf{28.03} & \textbf{46.18} & \textbf{25.64} \\
    \midrule
    Random Action & UM & 202.98 & 0.00 & 0.00 & 0.00 & 158.46 & 0.00 & 0.00 & 0.00 & 265.88 & 0.00 & 0.00 & 0.00 \\
    Fixed Action & UM & 180.47 & 0.52 & 2.61 & 0.39 & 132.89 & 0.89 & 4.28 & 0.67 & 247.72 & 0.00 & 0.25 & 0.00 \\
    CMA~\citep{wang2024openuav} & UM & 141.68 & 2.30 & 10.02 & 2.16 & \textbf{102.29} & 3.57 & 14.26 & 3.33 & 197.35 & 0.50 & 4.03 & 0.50 \\
    TravelUAV~\citep{wang2024openuav} & UM & \underline{138.80} & \underline{4.18} & \textbf{20.77} & \underline{3.84} & 102.94 & \underline{4.63} & \textbf{22.82} & \underline{4.24} & \underline{189.46} & \underline{3.53} & \textbf{17.88} & \underline{3.28} \\
    \rowcolor{myblue}
    \textbf{\ours (Four views)} & UM & \textbf{125.10} & \textbf{6.30} & \underline{18.95} & \textbf{5.68} & \underline{102.41} & \textbf{6.77} & \underline{20.07} & \textbf{6.04} & \textbf{170.58} & \textbf{5.36} & \underline{15.71} & \textbf{4.97} \\
    \bottomrule
\end{tabular}
}
\end{table}

\begin{wrapfigure}{r}{0.5\linewidth}
  \centering
    \includegraphics[width=\linewidth]{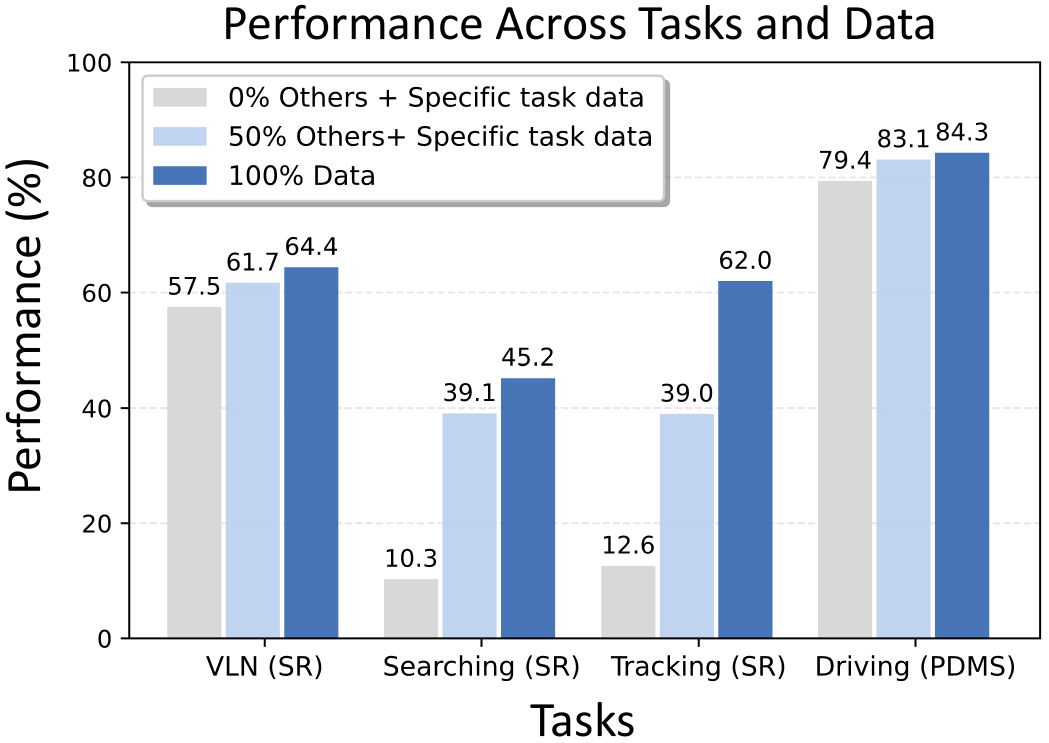}
  \caption{\textbf{Ablation study on the training of multiple navigation tasks.} We report the performance of different training data combinations (specific task data only, specific task data with 50\% other data, and specific task data with 100\% other data). $\dagger$ Searching is tested in a zero-shot manner.}
  \label{fig:ablation_task}
\end{wrapfigure}

\textbf{VLN: Performance on OpenUAV}~\citep{wang2024openuav}. We report the performance of our method in a challenging UAV scenario (Table~\ref{tab:openuav_combined_results}), which requires the UAV to follow natural language instructions and execute long-horizon trajectories (averaging 200 meters) to reach described targets in outdoor environments. Note that our method uses trajectories directly collected from the TravelUAV~\citep{wang2024openuav} training split (mimicking ground truth trajectories), as no strong baseline was available to collect expert trajectories as was done for the ObjectNav data collection. Despite this, our approach achieves state-of-the-art performance compared to prior UAV-specific baselines such as TravelUAV, without relying on downward-facing cameras as used in those methods (we plan to incorporate additional degrees of freedom in camera configurations in future work). This clearly demonstrates the effectiveness of our approach and the benefits of learning from diverse navigation tasks (Figure~\ref{fig:ablation_task}).

However, we observe that all methods perform poorly on the Unseen-Map split, which requires an average traversal of 300 meters through complex neighborhoods to reach unseen targets. This is because the unseen split demands more advanced navigation capabilities, such as efficient exploration of large-scale environments, which in turn relies on higher-quality UAV data.

% \input{iclr/table/openuav_seen}

% \input{iclr/table/oepnuav_unseen}

% \subsubsection{Searching}

\begin{table}[!t]
\centering
\caption{\textbf{Object goal navigation.} Comparison on HM3D-OVON~\cite{yokoyama2024hm3d}. $^*:$ denotes zero-shot evaluation. We report the performence of our method on egocentric and four-view settings. The \textbf{best} and the \underline{second best} results are denoted by \textbf{bold} and \underline{underline}.}
% \vspace{-5pt}
% \resizebox{\linewidth}{!}{
\scalebox{0.8}{
\setlength{\tabcolsep}{3mm}{
\begin{tabular}{lcccccc}
\hline
\multirow{2}{*}{Method} & \multicolumn{2}{c}{VAL SEEN}   & \multicolumn{2}{c}{\makecell{VAL SEEN \\ SYNONYMS}} & \multicolumn{2}{c}{VAL UNSEEN} \\
% \cline{2-7}
& SR$\uparrow$ & SPL$\uparrow$ & SR$\uparrow$ & SPL$\uparrow$ & SR$\uparrow$ & SPL$\uparrow$ \\ \midrule

BC & 11.1 & 4.5 & 9.9 & 3.8 & 5.4 & 1.9 \\
DAgger & 11.1 & 4.5 & 9.9 & 3.8 & 5.4 & 1.9 \\
RL & 18.1 & 9.4 & 15.0 & 7.4 & 10.2 & 4.7 \\
BCRL & 39.2 & 18.7 & 27.8 & 11.7 & 18.6 & 7.5 \\
DAgRL & 41.3 & 21.2 & 29.4 & 14.4 & 18.3 & 7.9 \\
VLFM$^*$~\citep{yokoyama2024vlfm} & 35.2 & 18.6 & 32.4& 17.3 & 35.2 & 19.6 \\
DAgRL+OD~\citep{yokoyama2024hm3d} & 38.5 & 21.1 & 39.0 & 21.4 & 37.1 & 19.8 \\
Uni-NaVid*~\citep{zhang2024uni} & \underline{41.3} & 21.1 & 43.9 & 21.8 & 39.5 & 19.8 \\
MTU3D~\citep{zhu2025mtu} & \textbf{55.0} & 23.6 & \underline{45.0} & 14.7 & 40.8 & 12.1 \\
\rowcolor{myblue}
\textbf{\ours* (Single view)} &  37.7 & \underline{25.5} &  43.3 &  \underline{29.9} & \underline{43.6} & \underline{31.3} \\ 
% \midrule
\rowcolor{myblue}
\textbf{\ours* (Four views)} & 40.1 & \textbf{27.1} & \textbf{45.4} & \textbf{32.6} & \textbf{45.2} & \textbf{31.9} \\
% \hline
\bottomrule
\end{tabular}
}
}
% }

% \vspace{-5mm}
\label{tab:openvocab-objnav}
\end{table} 


\textbf{Searching: Performence on OVON}~\citep{yokoyama2024hm3d}. Following prior work~\citep{zhang2024uni,zhu2025mtu}, we evaluate search capability on an open-vocabulary benchmark under a zero-shot setting. The results are presented in Table~\ref{tab:openvocab-objnav}, which includes performance for both single-camera and four-camera configurations.
Under the single-camera setting, our method achieves performance comparable to that of the state-of-the-art (SOTA) approach~\citep{zhu2025mtu} on both the VAL SEEN and VAL SEEN SYNONYMS splits in a zero-shot evaluation setting. On the more challenging VAL UNSEEN split, our method outperforms the SOTA method, improving the success rate (SR) from $40.8\%$ to $43.6\%$.
Furthermore, when transitioning from the single-camera to the four-camera setting, we observe consistent improvements across all splits and metrics. Notably, our model was trained only on single-camera search samples, demonstrating that co-tuning across different camera configurations enhances generalization to varied camera setups.

% \subsubsection{Tracking}
\begin{table}[]
        \caption{\textbf{Performance on EVT-Bench}. $\dag$: Uses GroundingDINO~\citep{liu2023grounding} as the open-vocabulary detector. $^\ddag$: Uses SoM~\citep{yang2023set}+GPT-4o~\citep{openai2024introducing} as the visual foundation model. The \textbf{best} and the \underline{second best} results are denoted by \textbf{bold} and \underline{underline}.}
        \centering
        \resizebox{0.6\textwidth}{!}{
        \begin{tabular}{lcccc}
            \toprule
            \multirow{2}{*}{Method}& \multicolumn{2}{c}{Single Target} & \multicolumn{2}{c}{Distracted Target}  \\
            & SR$\uparrow$ & TR$\uparrow$  & SR$\uparrow$ & TR$\uparrow$   \\
            \midrule
            IBVS$\dag$~\citep{gupta2016novel} & 42.9 & 56.2 & 10.6 & 28.4  \\
            PoliFormer$\dag$~\citep{zeng2024poliformer}  & 4.67 & 15.5 & 2.62 & 13.2 \\
            EVT~\citep{zhong2024empowering} & 24.4 & 39.1 & 3.23 & 11.2 \\
            EVT$\ddag$~\citep{zhong2024empowering} & 32.5 & 49.9 & 15.7 & 35.7  \\
            Uni-NaVid~\citep{zhang2024uni} & 25.7 & 39.5 & 11.3 & 27.4  \\
            TrackVLA~\citep{wang2025trackvla} & \underline{85.1} & 78.6 & 57.6 & 63.2  \\
            \rowcolor{myblue}
            \textbf{\ours (Single view)} & 85.0 & \underline{80.5} & \underline{61.4} & \textbf{68.2}   \\
            % \midrule
            \rowcolor{myblue}
            \textbf{\ours (Four views)} & \textbf{88.4} & \textbf{80.7} & \textbf{62.0} & \underline{67.9}   \\
            \bottomrule
        \end{tabular}
        }
        \label{tab:evt-bench}
\end{table}
\textbf{Tracking: Performance in the EVT-Bench}~\citep{wang2025trackvla}. We evaluate our method on EVT-Bench (including both the Single Target and Distracted Target splits) under both single-view and four-view camera settings (Table~\ref{tab:evt-bench}). Note that our model is trained only on the single-view setting and evaluated on the four-view setting in a zero-shot manner. Our results demonstrate that the proposed method achieves state-of-the-art (SOTA) performance under the single-view setting, outperforming the previous baseline, TrackVLA~\citep{wang2025trackvla}, which was specifically fine-tuned on tracking data. Furthermore, when the camera setup is increased from single-view to four-view (in a zero-shot manner), our method continues to improve its performance. However, compared to the improvement observed in VLN (a $6.8\%\uparrow$ in SR on VLN-CE RxR), the gains here are relatively modest ($0.6\%\uparrow$ in SR). We attribute this to the fact that most targets in EVT-Bench are spawned in front of the robot, a key assumption of this benchmark. We plan to further investigate this issue through both simulation and methodological enhancements, such as incorporating randomly positioned surrounding targets in future work.

\begin{table}[t]
\centering
\caption{\textbf{Comparison on planning-oriented NAVSIM \texttt{navtest} split with closed-loop metrics.} $\mathcal{V}_{8192}$ denotes 8192 anchors. The \textbf{best} and the \underline{second best} results are denoted by \textbf{bold} and \underline{underline}.
}
\label{tab:navsim_results_final}

\resizebox{\columnwidth}{!}{
\begin{tabular}{l ccc c ccccc|c}
    \toprule
    \multirow{2}{*}{Method} & \multicolumn{3}{c}{Observation \& Structure} & & \multicolumn{6}{c}{Metrics} \\
    \cmidrule(lr){2-4} \cmidrule(lr){6-11}
    & Camera & Lidar & VLM-Based & & NC $\uparrow$ &DAC $\uparrow$ & TTC $\uparrow$& Comf. $\uparrow$ & EP $\uparrow$ & PDMS $\uparrow$  \\
    \midrule
    Human & - & - & - & & 100 & 100 & 100 & 99.9 & 87.5 & 94.8 \\
    % \midrule
    Constant Velocity & - & - & - & & 69.9 & 58.8 & 49.3 & 100 & 49.3 & 21.6 \\
    Ego Status MLP & - & - & - & & 93.0 & 77.3 & 83.6 & 100 & 62.8 & 65.6 \\
    \midrule
    LTF~\citep{chitta2022transfuser} & \checkmark & \checkmark & - & & 97.4 & 92.8 & 92.4 & \textbf{100} & 79.0 & 83.8 \\
    Transfuser~\citep{chitta2022transfuser} & \checkmark & \checkmark & - & & 97.7 & 92.8 & 92.8 & \textbf{100} & 79.2 & 84.0 \\
    VADv2-$\mathcal{V}_{8192}$~\citep{chen2024vadv2} & \checkmark & \checkmark & - & & 97.2 & 89.1 & 91.6 & \textbf{100} & 76.0 & 80.9 \\
    Hydra-MDP-$\mathcal{V}_{8192}$~\citep{li2024hydra} & \checkmark & \checkmark & - & & 97.9 & 91.7 & 92.9 & \textbf{100} & 77.6 & 83.0 \\
    DiffusionDrive~\citep{liao2024diffusiondrive} & \checkmark & \checkmark & - & & \textbf{98.2}  & \textbf{96.2}  & \underline{94.7}  & \textbf{100}  & \textbf{82.2}  & \textbf{88.1}\\
    DRAMA~\citep{yuan2024drama} & \checkmark & \checkmark & \checkmark & & \underline{98.0} & \underline{93.1} & \textbf{94.8} & \textbf{100} & \underline{80.1} & \underline{85.5} \\
    \midrule
    UniAD~\citep{related_work_driving_UniAD} & \checkmark & - & - & & 97.8 & 91.9 & 92.9 & \textbf{100} & 78.8 & 83.4 \\
    PARA-Drive~\citep{related_work_driving_PARA_Drive} & \checkmark & - & - & & \underline{97.9} & 92.4 & \underline{93.0} & 99.8 & 79.3 & 84.0 \\
    LAW~\citep{LAW} & \checkmark & - & - & & 96.4  & \textbf{95.4}  & 88.7  & 99.9  & \textbf{81.7}  & \textbf{84.6}\\
    DrivingGPT~\citep{chen2024drivinggpt} & \checkmark & - & \checkmark & & \textbf{98.9}  & 90.7  & \textbf{94.9}  & 95.6  & \underline{79.7}  & 82.4\\
    % \midrule
    \rowcolor{myblue}
    \textbf{\ours (Eight views)} & \checkmark & - & \checkmark & & 97.7  & \underline{93.5}  & 92.3  & \textbf{100}  & 79.6  & \underline{84.3}\\
    \bottomrule
\end{tabular}
}
\end{table}

\textbf{Autonoums Drivning: Perfomrence on NAVSIM}~\citep{Dauner2024navsim} and \textbf{nuScenes}~\citep{caesar2020nuscenes}. We conduct experiments to evaluate our method under six-view and eight-view settings (without fine-tuning for specific configurations). Results on NAVSIM and nuScenes are reported in Table~\ref{tab:navsim_results_final} and Appendix Table~\ref{tab:nuscene_results_hierarchical}, respectively. We observe that our method achieves performance comparable to SOTA methods on both benchmarks, without explicitly modeling driving-related information such as lane markings, nearby vehicles, or other contextual elements. We believe our approach can be further improved by incorporating scene descriptions as prompts, similar to other baseline methods. We are also interested in evaluating this model in closed-loop autonomous driving simulators such as \citep{dosovitskiy2017carla}.

\begin{figure}[t]
  \centering
    \includegraphics[width=\linewidth]{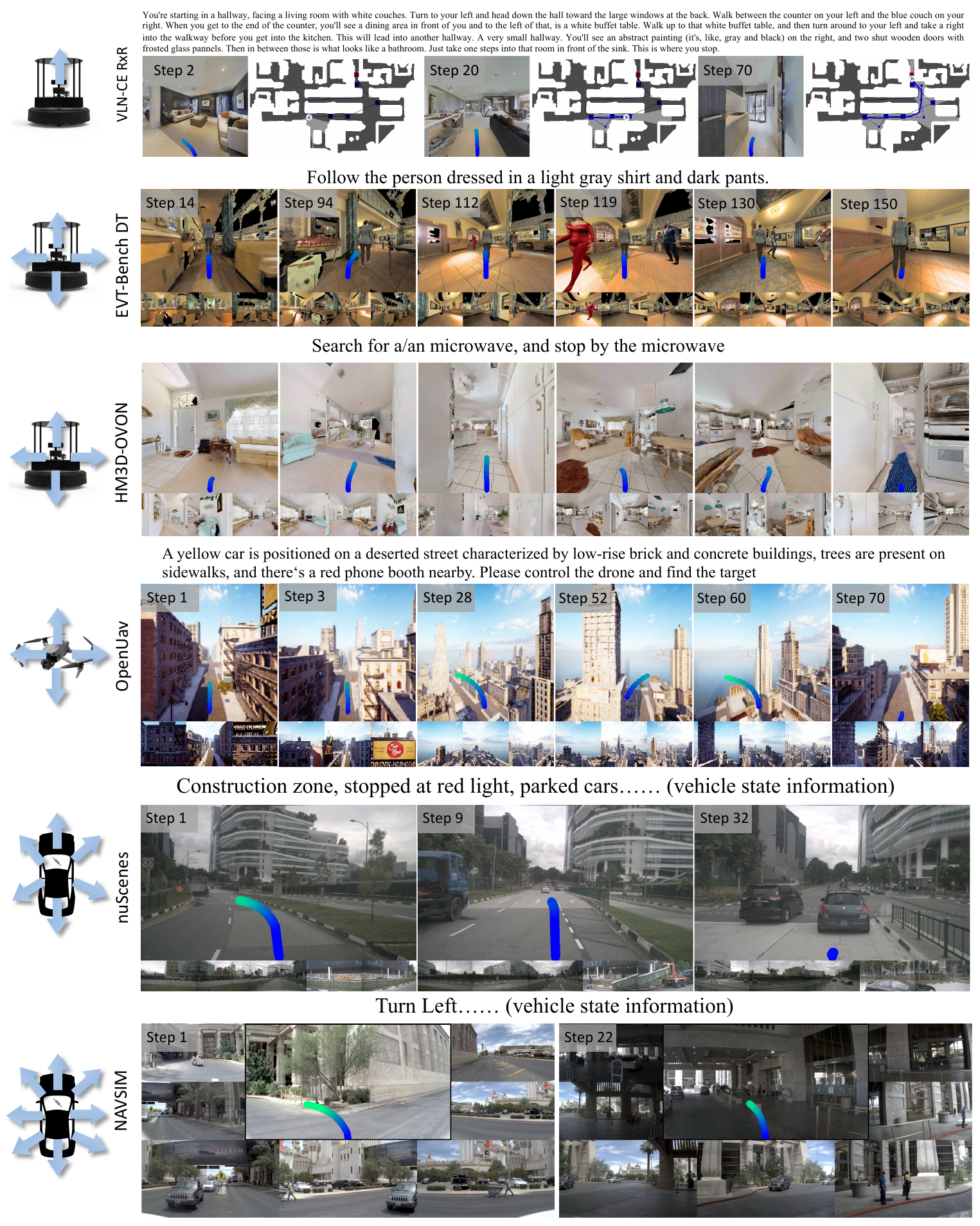}
  \caption{\textbf{Visualization of perfomrence on benchmarks.} We report visual results of \ours on VLN-CE RxR (single-view), EVT-Bench Distracted Targets (four-view), OpenUAV (four-view), NeuScenes (six-view), OpenScenses (Eight-view).}
  \label{fig:synthetic_bench_vis}
  \vspace{-5mm}
\end{figure}
\textbf{Visualization of benchmark results.} 
We provide visual results of \ours on above benchmarks in Figure~\ref{fig:synthetic_bench_vis}, where we plot the predicted trajcotry, camera views, correspongding instructions.

\subsection{Real-World Results}
\label{sec:real-world-exp}

\begin{figure}[ht]
  \centering
    \includegraphics[width=1.0\linewidth]{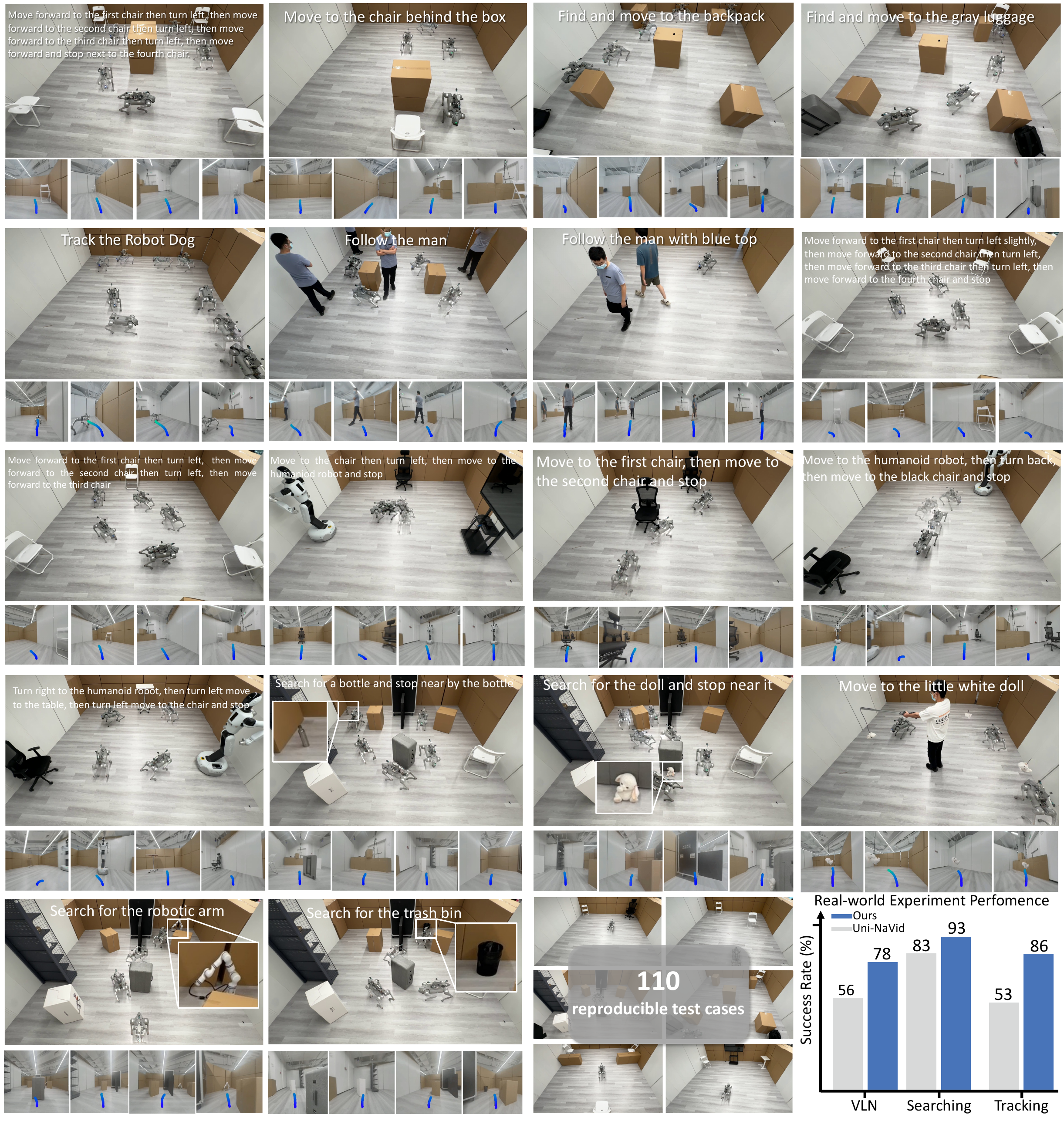}
  \caption{\textbf{Real-world experiments.} We report both the qualitive and quantitive results of \ours on complex seniors among different navigation cabability. }
  \label{fig:real_world_exp}
\end{figure}

\textbf{Real-world performence on 110 reproducible test cases.} To evaluate the real-world performance of our method, we designed a series of navigation test cases with different capabilities (including 50 VLN samples, 30 search samples, and 30 tracking samples). Specifically, we constructed a $5\text{m} \times 5\text{m}$ space and recorded the locations of the robot, obstacles, and targets for each test case. We report both qualitative and quantitative results of \ours in complex scenarios across these navigation capabilities. The results are presented in Figure~\ref{fig:real_world_exp}. Our findings indicate that \ours demonstrates strong real-world performance: it correctly understands the surrounding environment and plans appropriate trajectories to accomplish the task. Moreover, compared to the strong baseline Uni-NaVid \citep{zhang2024uni}, our method exhibits significant improvements across both tasks, demonstrating its superior performance in real-world environments.

\begin{figure}[t]
  \centering
    \includegraphics[width=\linewidth]{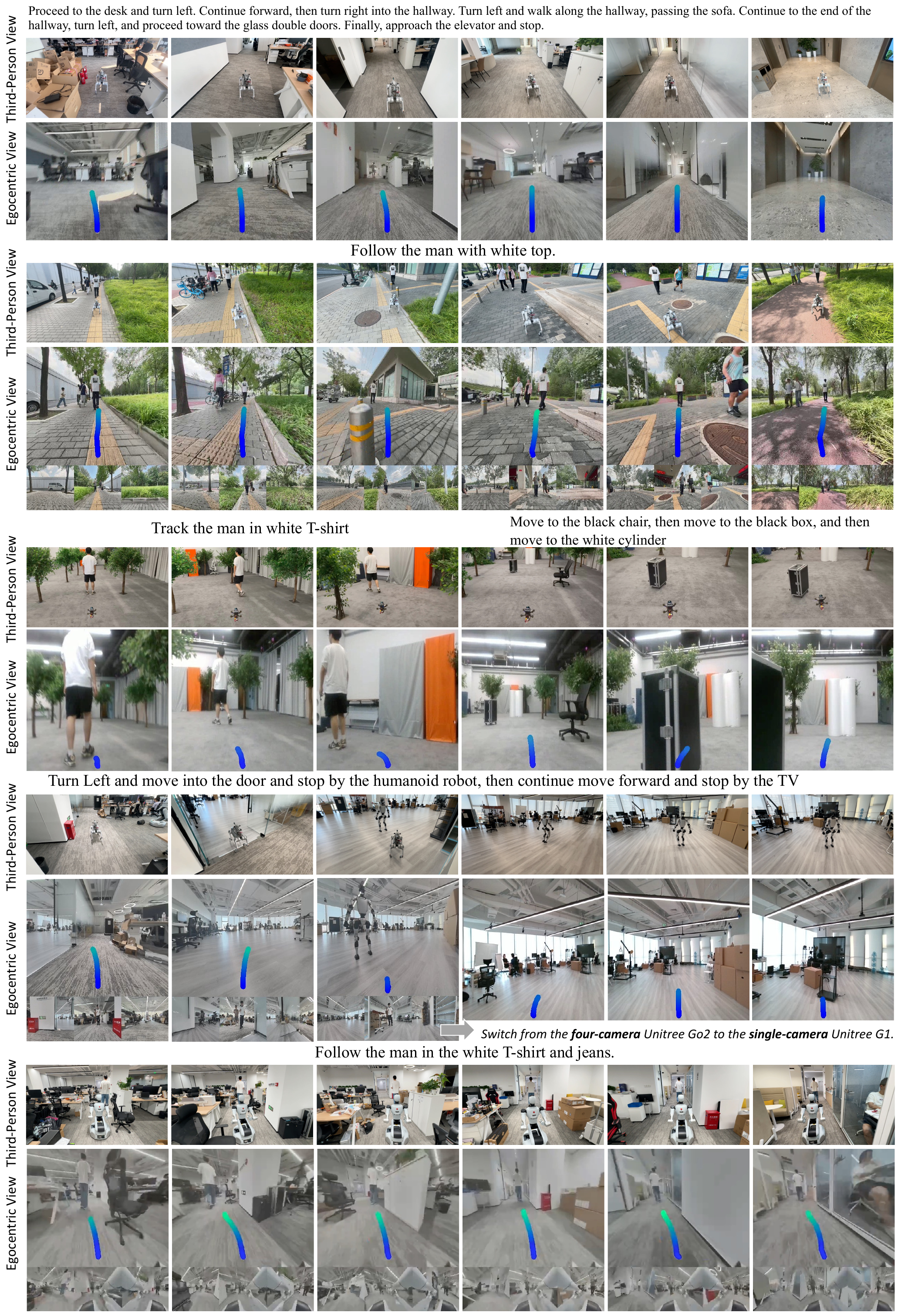}
  \caption{Visualization of real-world experiments on cross-task and cross-embodiment settings.}
  \label{fig:real_cross}
  \vspace{-5mm}
\end{figure}

\textbf{Visual results of challenging cross-task and cross-emdbodiement real-world experiments.}
We also conduct extensive experiments on more challenging scenarios with different embodiments (quadruped robots, humanoids, drones, and wheeled robots). The results are shown in Figure~\ref{fig:real_cross}, where we find that our method can handle complicated real-world environments and fulfill long-horizon instructions. We encourage readers to view our accompanying videos for a more intuitive demonstration.

% \subsubsection{Cross-embodiments}

\clearpage

\subsection{Ablation Study}

% \begin{figure}[ht]
%   \centering
%     \includegraphics[width=0.5\linewidth]{iclr/figure/ablation_samples_num.pdf}
%   \caption{\textbf{Ablation study on the training of multiple navigation tasks.} We report the performance of different training data combinations (specific task data only, specific task data with 50\% other data, and specific task data with 100\% other data). $\dagger$ Searching is tested in a zero-shot manner.}
%   \label{fig:ablation_task}
% \end{figure} 

% \begin{wrapfigure}{r}{0.5\linewidth}
%   \centering
%     \includegraphics[width=\linewidth]{iclr/figure/cached_feature.png}
%   \caption{Offline Visual Feature Cached.}
% \end{wrapfigure}

\textbf{Synergy of training on multiple navigation tasks.}
We investigate the synergistic effects of multi-navigation task training by comparing the performance of single-task training with co-tuning that incorporates additional data from other navigation tasks (at $50\%$ and $100\%$ ratios). Here, "VLN" refers to the VLN-CE RxR four-view setting, "Searching" refers to the OVON four-view setting (evaluated in a zero-shot manner), "Tracking" refers to the EVT-Bench four-view setting, and "Driving" refers to the NavSIM eight-view setting. We observe that co-tuning with data from diverse navigation tasks leads to consistent performance improvements across all tasks (from $50\%$ to $100\%$ data ratios). Notably, Searching (improving from $10.3\%$ to $45.2\%$) and Tracking (improving from $12.6\%$ to $62.0\%$) exhibit the most significant gains. We attribute these improvements to the discrepancy between their training conditions (primarily single-view and closed-set target categories) and the evaluation settings, which are multi-view and open-vocabulary. By learning from multi-view data with diverse targets sourced from other tasks, both Searching and Tracking enhance their multi-view and open-set navigation capabilities, resulting in a substantial performance boost. These results suggest that training across multiple navigation tasks helps mitigate overfitting to task-specific navigation patterns.

\begin{figure}[ht]
  \centering
    \includegraphics[width=0.5\linewidth]{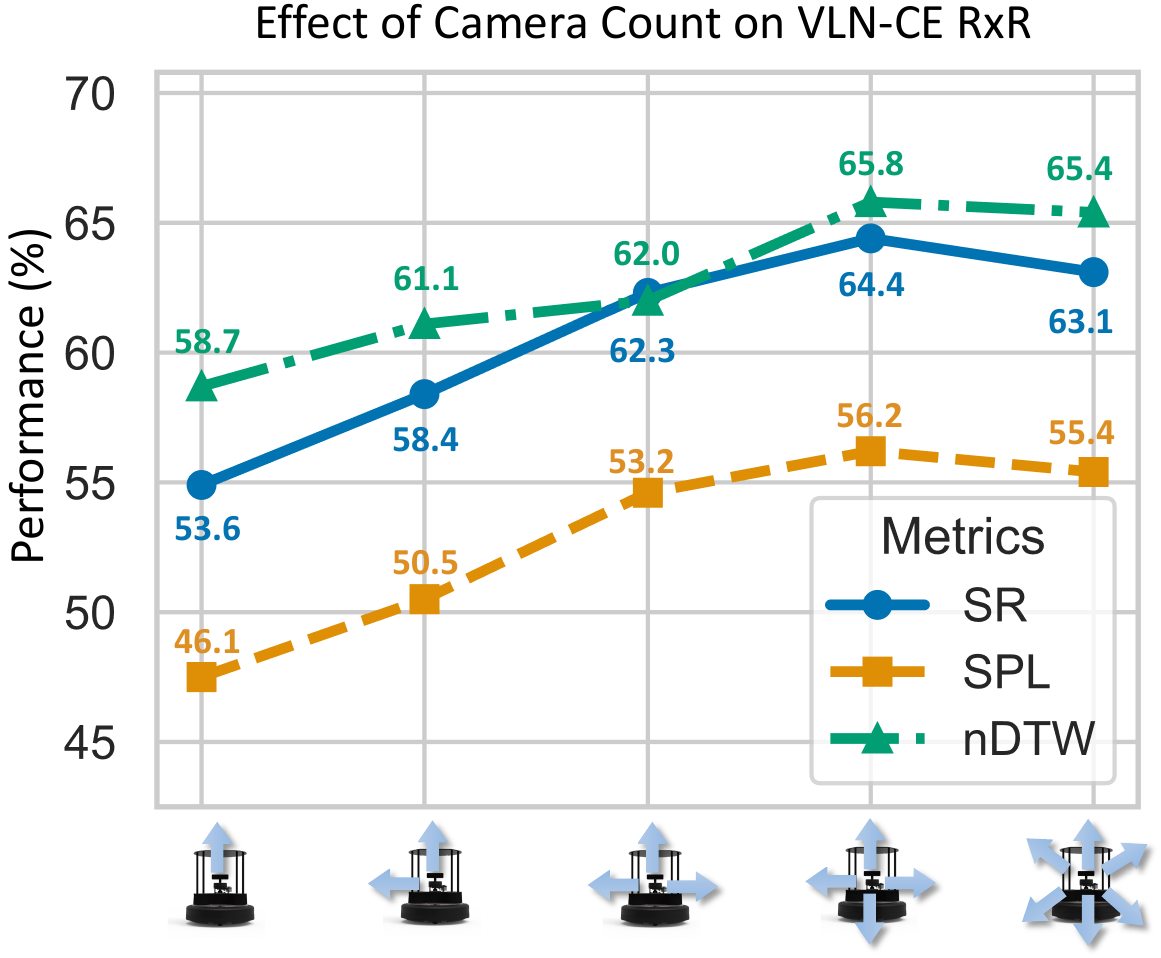}
  \caption{\textbf{Ablation study on the number of cameras in VLN-CE RxR.} We report the performance under five different camera configurations (from left to right: one-, two-, three-, four-, and six-camera settings), with same token budget ($B=2048$).}
  \label{fig:ablation_camera_num}
\end{figure}
\textbf{Performance on differnt number of cameras.} We evaluate the effectiveness of incorporating additional cameras in navigation tasks on VLN-CE RxR, a benchmark that offers a relatively comprehensive suite of vision-language navigation challenges. The results are presented in Table~\ref{fig:ablation_camera_num}, which compares configurations of one, two, three, four, and six cameras mounted around the robot to achieve a wider field of view. We observe consistent performance improvements when increasing the number of cameras from one to four, validating that enhanced environmental observations contribute positively to navigation performance. Notably, however, expanding to six cameras leads to a slight degradation in performance. We attribute this to the fact that six cameras do not provide substantially more observational coverage compared to four cameras, while the increased number of view tokens reduces the capacity available for encoding historical frames (Equation~\ref{eq:bats}). This weaks the alignment between the navigation history and the instruction. We suggest that this issue could be mitigated by adopting an adaptive multi-view token encoding strategy. To maintain coherence in the current work, we leave this exploration for future research.

\begin{table}[!t]
\centering

% \vspace{-5pt}
% \resizebox{\linewidth}{!}{
\scalebox{0.85}{
\setlength{\tabcolsep}{0.85mm}{
% \usepackage{multirow}
% \begin{tabular}{l|cccc}
% \hline
% \multicolumn{3}{l|}{Type}             & VLN (SR$\uparrow$) & ObjNav (SR$\uparrow$) & EQA (ACC$\uparrow$) & Follow (SR$\uparrow$) \\ 

\begin{tabular}{l cccc}
\toprule
\multirow{2}{*}{Type}   & \multicolumn{4}{c}{RxR Val-Unseen} \\
\cmidrule(lr){2-5} 
 & NE $\downarrow$ & SR $\uparrow$ & SPL $\uparrow$ & nDTW $\uparrow$ \\
\midrule
% \hline
$B=1024$, Uniform Sampling*       & 5.33 & 59.7 & 49.6 & 57.9 \\ 
$B=1024$, Linear Probability Sampling      & 5.28 & 61.2 & 50.9 & 58.9 \\
$B=1024$, Budget-Aware Temporal Sampling    & 4.98 & 62.5 & 53.9 & 64.1 \\
$B=2048$, Token Merging~\citep{zhang2024uni} & 5.01 & 63.2 & 54.9 & 64.4 \\
$B=2048$, Uniform Sampling*     & 4.90 & 62.4 & 54.0 & 63.9 \\ 
$B=2048$, Linear Probability Sampling       & 4.89 & 63.0 & 54.6 & 64.8 \\
% Budget-Aware Temporal Sampling ($B=1600$)    & 4.84 & 63.5 & 55.1 & 65.0 \\
$B=2048$, Budget-Aware Temporal Sampling    & \textbf{4.74} & \textbf{64.4} & \textbf{56.2} & \textbf{65.8} \\ \midrule
Viewpoint-history postional embedding$^\dagger$      & 6.27 & 52.3 & 46.3 & 58.7 \\
Individual Learned Special Toekns$^\ddagger$       & 5.52 & 59.1 & 52.0 & 59.6 \\
Handcraft Toekns (Equ.~\ref{equ:TVI_Tokes} w.o $\mathcal{P}_{\text{angle}/\text{time}}$) & 6.06      & 53.6       & 46.1       & 58.0         \\
Temporal-Viewpoint Indicator Tokens (Equ.~\ref{equ:TVI_Tokes}) & \textbf{4.74} & \textbf{64.4} & \textbf{56.2} & \textbf{65.8}     \\
\bottomrule
\end{tabular}
}
}
\caption{\textbf{Ablation study on history token organization strategies and identity tokens.} For both linear probability sampling and uniform sampling, all methods sample from a base set of 2048 tokens. *Uniform sampling refers to the technique used in \citep{cheng2024navila}. $\dagger$We employ the common positional encoding technique from \cite{chen2021hamt}, which is widely adopted in traditional VLN methods. $^\ddagger$We introduce predefined learnable special tokens for each viewpoint in the datasets. }
\label{tab:ablation_study_BATS_TVI}
% \vspace{-4mm}

\end{table}

% \begin{table}[!t]
% \centering

% % \vspace{-5pt}
% % \resizebox{\linewidth}{!}{
% \scalebox{0.7}{
% \setlength{\tabcolsep}{0.7mm}{
% % Please add the following required packages to your document preamble:
% % \usepackage{multirow}
% \begin{tabular}{ccc|cccc}
% \hline
% Curr.      & Short.     & Long.      & VLN (SR) & ObjectNav (SR) & EQA (ACC) & Following (SR) \\ \hline
% \checkmark &            &  & 9.1 (81.3\%$\downarrow$)  & 44.3 (39.8\%$\downarrow$) & 32.5 (31.2\%$\downarrow$) & 56.3 (8\%$\downarrow$)   \\
% \checkmark & \checkmark &  & 39.7 (18.4\%$\downarrow$) & 67.8 (8.0\%$\downarrow$)  & 44.1 (6.7\%$\downarrow$)  & 59.7 (2.4\%$\downarrow$) \\
% \checkmark & \checkmark & \checkmark & 48.7     & 73.7           & 47.3      & 61.2           \\ \hline
% \end{tabular}
% }
% }
% \caption{\textbf{Ablation study on memory strategy.}}
% \label{tab:ablation-memory}
% \end{table} 

\textbf{Effectiveness of BATS and TVI tokens.} We conduct ablation studies to evaluate the effectiveness of our key designs, including the history token organization strategy and visual-temporal history modeling. The experiments are conducted on the VLN-CE RxR four-camera setting, and the results are presented in Table~\ref{tab:ablation_study_BATS_TVI}. We test different token strategies under different token budgets (1024 or 2048) and find that BATS outperforms other strategies in both settings. In particular, on the nDTW metric, which directly measures the alignment between the navigation trajectory and the ground-truth trajectory, BATS exhibits only a minor performance drop ($1.4\%\downarrow$), compared to uniform sampling~\citep{cheng2024navila} ($6.0\%\downarrow$) and linear probability sampling ($5.2\%\downarrow$).
Furthermore, we compare TVI tokens with other common alternatives (individually learned special tokens and handcrafted tokens as in Equation~\ref{equ:TVI_Tokes} without $\mathcal{P}_{\text{angle}/\text{time}}$) and find that TVI tokens achieve significantly better performance. As illustrated in Figure~\ref{fig:tvi_tokens}, we attribute this improvement to the well-learned temporal and viewpoint information.
In contrast, compared to the common history-viewpoint positional embedding method~\citep{chen2021hamt}, we observe a noticeable performance drop. We believe that this is due to the additional embedding components of visual tokens, which may increase the complexity of representation learning for the LLM. These results suggest that the community may need new techniques to effectively incorporate viewpoint and temporal information and we believe TVI tokens represent a promising starting point.

% \haoran{cite the tables} \haoran{what is handcraft tokens} \haoran{baseline with no special tokens}

% \textbf{Effectivness of BATS tokens.}

% \section{Limitation}

\section{Disscusion and Conclusion}
In this work, we propose \ours, which aims to push the boundaries of navigation and explore the intelligence learned from cross-embodiment and cross-task navigation data. We introduce temporal-viewpoint indicator tokens to enhance the LLM's understanding of varying camera configurations and different horizons in navigation tasks, while also enabling co-training with navigation and question-answering data. Furthermore, we employ a token budget-aware temporal sampling strategy to balance navigation performance and efficiency, facilitating a unified approach to token sampling across diverse camera setups and task horizons. Extensive experiments on both public benchmarks and real-world environments demonstrate that \ours achieves impressive performance and shows strong potential for further improvement with more advanced techniques or higher-quality data.

We believe that \ours serves merely as a starting point toward a navigation foundation model. We hope that this work will attract greater attention to intelligence-centric navigation and inspire a new generation of techniques, datasets, and benchmarks.

% which could inspire many directions of thie filed, including \textit{benchmakrs}, \textit{task},

% to the navigation foundation model 

\bibliography{iclr2025_conference}
\bibliographystyle{iclr2025_conference}

\clearpage

\appendix
\section{Implemetation Detials}

\subsection{Action planning model}
\label{appendix:action}

\begin{wraptable}{r}{0.4\linewidth}
    \centering
    \vspace{-10pt}
    \resizebox{0.35\textwidth}{!}{
    \begin{tabular}{lcccc}
\toprule
Embodiements & x(m) & y(m) & z(m)& $\theta$(rad) \\
    \midrule
        Indoor robots* & 1.0 & 0.433 & -  & 2.09  \\
        UAV* & 7.93 & 3.19 &  7.85 & 1.04  \\
        Cars* & 50.8 & 14.9 & - & 1.52  \\
    \bottomrule
\end{tabular}}
\caption{Scaling factors of different dimesiong of predicted tracjtort of different embodiements.}
\label{tab:alpha_task}
\end{wraptable}

% \begin{wraptable}{r}{0.5\textwidth}
%     \centering
%     \vspace{-10pt}
%     \resizebox{0.5\textwidth}{!}{
%         \begin{tabular}{lccccc}
%         \toprule
%         Model & Params. & SR$\uparrow$ & TR$\uparrow$ & CR$\downarrow$ & time(ms) $\downarrow$ \\
%         \midrule
%         Autoregressive  & 131M  & 42.6  &  56.9    &  11.7    &  460  \\
%         MLP (3-Layers)  & 7M  & 45.8  &  59.9    &  10.1    &   \textbf{0.5}   \\
%         MLP (6-Layers)  & 89M  & 52.7  &  61.9    &  9.42   &   0.8  \\
%         DP-Base & 89M &  17.9  &  33.8    &  27.7    &   65   \\
%         Ours-Small &  13M    & 49.8 &   60.2   &  6.67   &  8    \\
%         \rowcolor{gray!20}
%         Ours-Base & 89M    & \textbf{57.6} &  \textbf{63.2}   &  \textbf{5.80}    & 13    \\
%         % Ours-Large &  562M    & 53.1 &   61.5   &  7.16    &   24   \\
%         \bottomrule
%         \end{tabular}}
%     \vspace{-8pt}    
%     \caption{Comparison of different action models.}
%     \vspace{-10pt}
%     \label{tab:ablation_action}
% \end{wraptable}
Due to the fact that different embodiments could have distinct trajectory scales. For instance, indoor robots often move on the scale of meters while cars move on the scale of dozens of meters. We normalize the predicted trajectory scaling across different embodiments to the range [-1,1] of all dimesions by multipy a scaling factor $\alpha_\text{task}$, as reported in Table~\ref{tab:alpha_task}. Note that the scaling factor is not derived from the absolute maximum value of each dimension; instead, we use the 99th percentile of each dimension to avoid the influence of outlier data.

\subsection{Detials of using BATS}
\label{appendix:BATS}
During navigation, initially when the number of visual tokens is within the token budget $B$, we retain all visual tokens. Once the visual tokens exceed the budget $B$, we employ BATS to sample tokens based on a forgetting curve (Sec.~\ref{sec:BATS}). In practice, we precompute $P(t,T)$ for a given token budget $B$ to accelerate this process. If the navigation task involves an exceptionally long horizon, such as thousands of steps (which rarely occurs), even using the minimum sampling probability $\epsilon$ may result in the visual tokens exceeding the token budget. In such cases, we simply remove the oldest frames.

\subsection{Navigation data preparation}
\label{appendix:nav_data}

\begin{figure}[H]
  \centering
    \includegraphics[width=\linewidth]{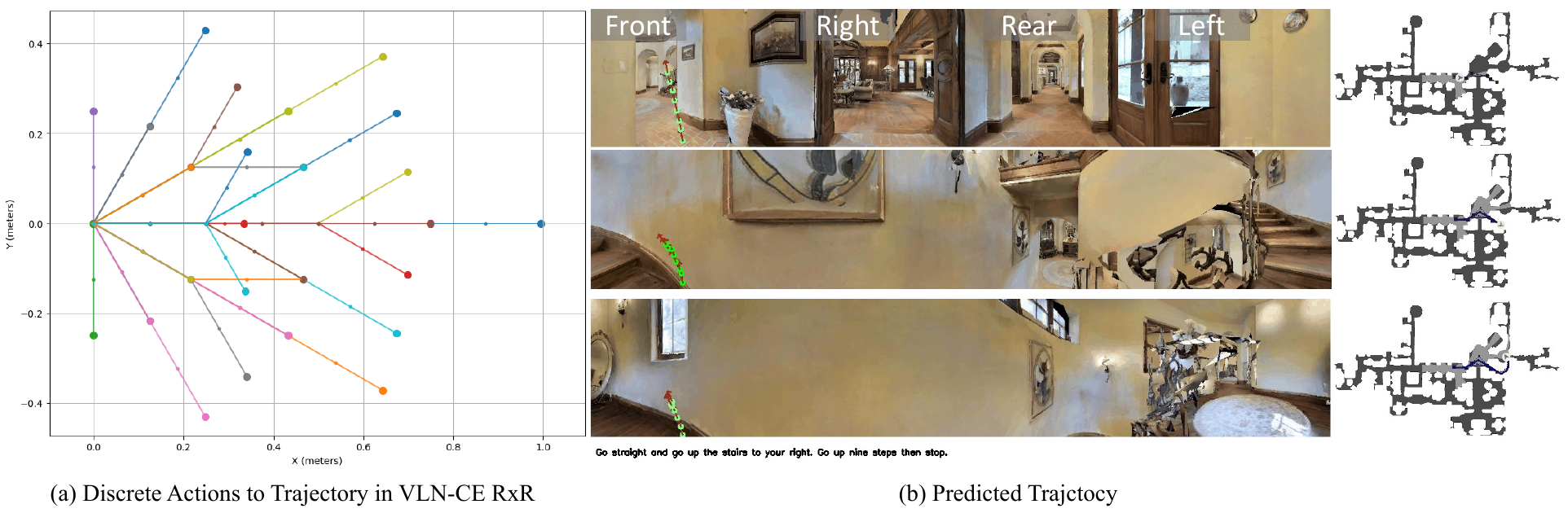}
  \caption{Visualization of the trajectory (VLN-CE RxR) for (a) training and (b) evaluation.}
  \label{fig:discrete_action_process}
\end{figure}

For navigation tasks built on the Habitat environment~\citep{habitat19iccv}, which utilizes low-level discrete actions such as \texttt{Move\_Forward}, \texttt{Turn\_Left}, \texttt{Turn\_Right}, and \texttt{Stop}. However, the definitions of these discrete actions vary slightly across different navigation tasks. For example, in VLN-CE R2R, \texttt{Turn\_Left} indicates a 15-degree turn, whereas in VLN-CE RxR and HM3D-ObjNav, it indicates a 30-degree turn. To unify all navigation tasks with discrete actions, we employ a simple strategy: we consider moving forward by 12.5 cm or turning by 15 degrees as an atomic operation. We then construct the trajectory based on the accumulation of these atomic operations. Although the resulting trajectory could be zigzag (Figure~\ref{fig:discrete_action_process}), after fine-tuning on all navigation datasets, we find that the predicted trajectory of our method is smooth and meaningfully directed toward the target.

\subsection{Real-world deployment system}

We regard our model as a general Visual-Language-Action (VLA) model capable of driving different embodiments to complete various navigation tasks. To achieve this, our model takes visual observations—obtained from one or more cameras—along with instructions, and directly predicts a trajectory. We then utilize off-the-shelf APIs (which may include Lidar or other sensors if necessary) specific to each embodiment to drive the robot along the predicted trajectory.
\begin{figure}[H]
  \centering
    \includegraphics[width=\linewidth]{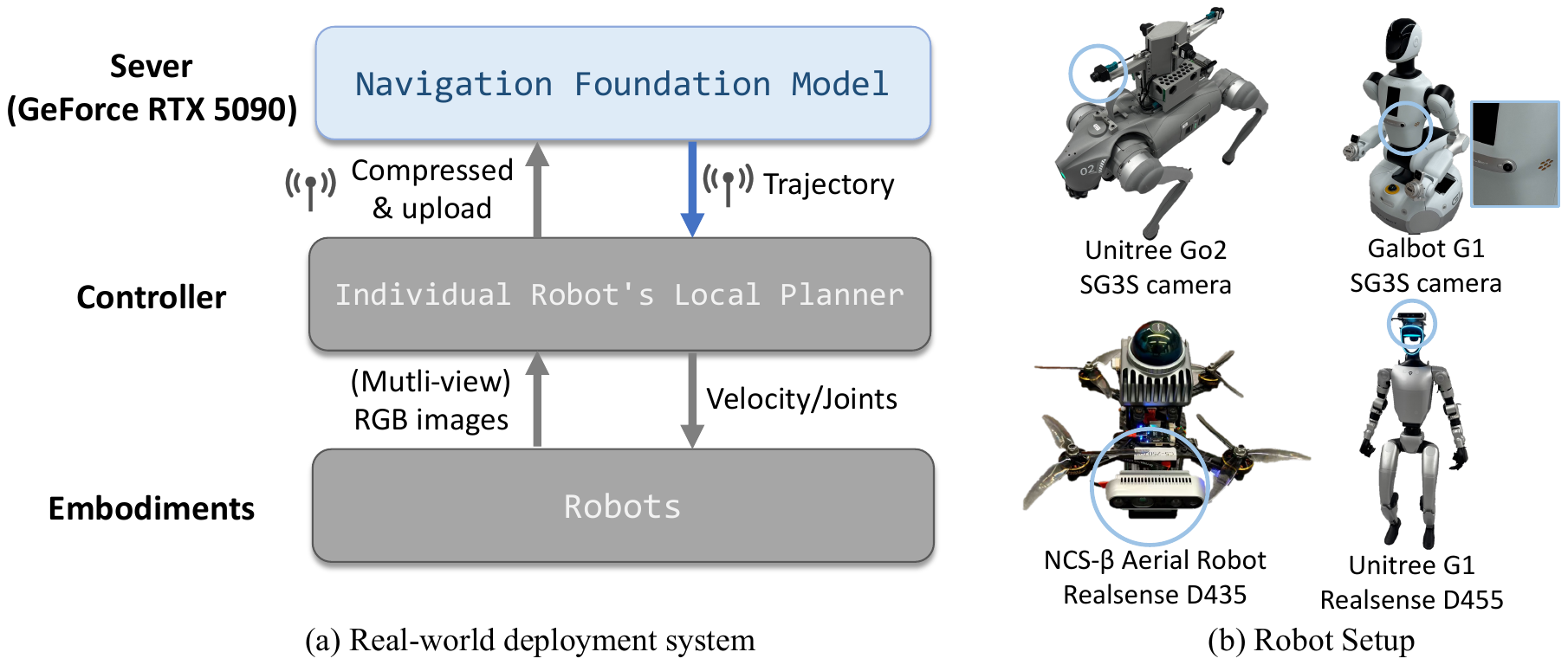}
  \caption{\textbf{Real-world deployment setup.} We provide the system architecture of our methods and the corresponding robots that were tested in the paper.}
  \label{fig:real_world_deploy}
\end{figure}

An illustration of our real-world system is provided in Figure~\ref{fig:real_world_deploy}. Specifically, we deploy our model on a remote server equipped with a GeForce RTX 5090 GPU and use the Internet for communication between the server and the client (which includes the controller and embodiments). Given a user instruction, the robots compress their current observations and transmit them to the server. The server then processes both the observations and the instruction to output a trajectory. This trajectory is subsequently processed by the local planner of each individual robot, which sends appropriate commands (\textit{e.g.}, velocity or joint controls) to drive the robot.

\section{Additional Experiment}

\subsection{Additional results on autonomous driving}

\begin{table}[H]
\centering
\caption{\textbf{Comparison on planning-oriented \texttt{nuScenes} dataset with open-loop metrics.} Metric calculation follows DiffusionDrive~\citep{related_work_driving_DiffusionDrive}. The \textbf{best} and the \underline{second best} results are denoted by \textbf{bold} and \underline{underline}.}
\label{tab:nuscene_results_hierarchical}

% \newcolumntype{g}{>{\columncolor{gray!30}}c} % This line is removed

\resizebox{\columnwidth}{!}{
% The 'g' specifiers are replaced with 'c'
\begin{tabular}{l  c c  ccc|c  ccc|c}
\toprule
\multirow{2}{*}{Method} &
\multicolumn{2}{c}{Observation \& Structure} &
\multicolumn{4}{c}{L2 (m) $\downarrow$} &
\multicolumn{4}{c}{Collision (\%) $\downarrow$} \\
\cmidrule(lr){2-3} \cmidrule(lr){4-7} \cmidrule(lr){8-11}
& Camera & VLM-Based & 1s & 2s & 3s & Avg. & 1s & 2s & 3s & Avg. \\
\midrule
ST-P3~\citep{related_work_driving_STP3} & \checkmark & - & 1.33 & 2.11 & 2.90 & 2.11 & 0.23 & 0.62 & 1.27 & 0.71 \\
UniAD~\citep{related_work_driving_UniAD} & \checkmark & - & 0.45 & 0.70 & 1.04 & 0.73 & 0.62 & 0.58 & 0.63 & 0.61 \\
VAD~\citep{related_work_driving_VAD} & \checkmark & - & 0.41 & 0.70 & 1.05 & 0.72 & 0.07 & 0.17 & 0.41 & 0.22 \\
SparseDrive~\citep{related_work_driving_SparseDrive} & \checkmark & - & 0.29 & 0.58 & 0.96 & 0.61 & \underline{0.01} & \textbf{0.05} & \underline{0.18} & \textbf{0.08} \\
DiffusionDrive~\citep{related_work_driving_DiffusionDrive} & \checkmark & - & 0.27 & 0.54 & 0.90 & 0.57 & 0.03 & \textbf{0.05} & \textbf{0.16} & \textbf{0.08} \\
% \midrule
DriveVLM~\citep{Driving_VLM_drivevlm} & \checkmark & \checkmark & 0.18 & 0.34 & 0.68 & 0.40 & 0.10 & 0.22 & 0.45 & 0.27 \\
EMMA~\citep{Driving_VLM_EMMA} & \checkmark & \checkmark & \textbf{0.14} & \textbf{0.29} & \textbf{0.54} & \textbf{0.32} & - & - & - & - \\
DME-Driver~\citep{Driving_VLM_DME-Driver} & \checkmark & \checkmark & 0.45 & 0.91 & 1.58 & 0.98 & 0.05 & 0.28 & 0.55 & 0.29 \\
Omni-Q~\citep{Driving_VLM_OmniDrive} & \checkmark & \checkmark & \textbf{0.14} & \textbf{0.29} & \underline{0.55} & \underline{0.33} & \textbf{0.00} & 0.13 & 0.78 & 0.30 \\
Omni-L~\citep{Driving_VLM_OmniDrive} & \checkmark & \checkmark & \underline{0.15} & 0.36 & 0.70 & 0.40 & 0.06 & 0.27 & 0.72 & 0.35 \\
ORION~\citep{Driving_VLM_orion} & \checkmark &\checkmark & 0.17 & \underline{0.31}& \underline{0.55} & 0.34& 0.05 & 0.25 & 0.80 & 0.37  \\
% \midrule
\textbf{\ours (Six views)} & \checkmark & \checkmark & 0.26 & 0.39 & 0.60 & 0.42 & 0.07 & \underline{0.11} & \underline{0.18} & \underline{0.12} \\
\bottomrule
\end{tabular}
}
\end{table}
We report the performance of our method on the autonomous driving benchmark nuScenes in Table~\ref{tab:nuscene_results_hierarchical}. We compare our method with strong baselines that are specifically designed for autonomous driving. Nevertheless, our method achieves comparable performance to these methods without explicitly modeling driving-related information.

% \textbf{Additional results on autonousms drving}

% \textbf{Real-world robot system setup}

% \textbf{Stop ciretia}

\end{document}